\documentclass[10pt,twocolumn,letterpaper]{article}
\usepackage[utf8]{inputenc}
\usepackage[font=normalfont,labelfont=bf]{caption}
\usepackage{lipsum}
\usepackage{float}
\usepackage{subfloat}
\usepackage{balance}
\usepackage{etoolbox}
\makeatletter
\providecommand{\institute}[1]{
  \apptocmd{\@author}{\end{tabular}
    \par
    \begin{tabular}[t]{c}
    #1}{}{}
}

\providecommand{\email}[1]{
  \apptocmd{\@author}{\end{tabular}
    \par
    \begin{tabular}[t]{c}
    #1}{}{}
}

\providecommand{\teaser}[1]{
  \apptocmd{\@author}{\end{tabular}
    \par
    \begin{tabular}[t]{c}
    #1}{}{}
}

\usepackage{cvpr}              

\usepackage{graphicx}
\usepackage{amsmath}
\usepackage{amssymb}
\usepackage{tabularx}
\usepackage{booktabs}
\usepackage{multirow}
\usepackage{subcaption}

\usepackage{makecell}
\usepackage{multirow}
\usepackage{arydshln}
\usepackage{booktabs}
\usepackage{colortbl}
\usepackage[ruled,vlined]{algorithm2e}
\usepackage{footnote}
\usepackage{url}

\SetKwSty{mykwsty}
\SetArgSty{textbf}
\SetFuncSty{myfuncsty}
\SetDataSty{myvarsty}
\usepackage[pagebackref,breaklinks,colorlinks]{hyperref}

\usepackage[capitalize]{cleveref}
\crefname{section}{Sec.}{Secs.}
\Crefname{section}{Section}{Sections}
\Crefname{table}{Table}{Tables}
\crefname{table}{Tab.}{Tabs.}

\def\cvprPaperID{*****} 
\def\confName{CVPR}
\def\confYear{2022}

\begin{document}

\title{SuperpixelGridCut, SuperpixelGridMean and SuperpixelGridMix \\Data Augmentation}

\author{Karim Hammoudi$^{1,2,*}$ \and Adnane Cabani$^{3,}$\thanks{Two authors contributed equally to this work (corresponding authors).} \and Bouthaina Slika$^{4}$ \and Halim Benhabiles$^{5}$ \and Fadi Dornaika$^{4,6}$ \and Mahmoud Melkemi$^{1,2}$}

\institute{$^{1}$Universit\'e de Haute-Alsace, IRIMAS \and $^{2}$Universit\'e de Strasbourg \and $^{3}$Normandie Universit\'e, ESIGELEC-IRSEEM \and $^{4}$University of the Basque Country \and $^{5}$JUNIA, CNRS-IEMN, Universit\'e de Lille \and $^{6}$IKERBASQUE}

\email{\small karim.hammoudi@uha.fr \and \small adnane.cabani@esigelec.fr \and \small bslika001@ikasle.ehu.eus \and \small halim.benhabiles@junia.com \and \small fadi.dornaika@ehu.eus \and \small mahmoud.melkemi@uha.fr}


\makeatletter
\let\@oldmaketitle\@maketitle
\renewcommand{\@maketitle}{\@oldmaketitle
\centering
\vspace*{-0.75cm}
\setlength{\tabcolsep}{0.1em} 
{\renewcommand{\arraystretch}{1.2}
\begin{tabular}{ccccc}
\includegraphics[height=2.55cm]{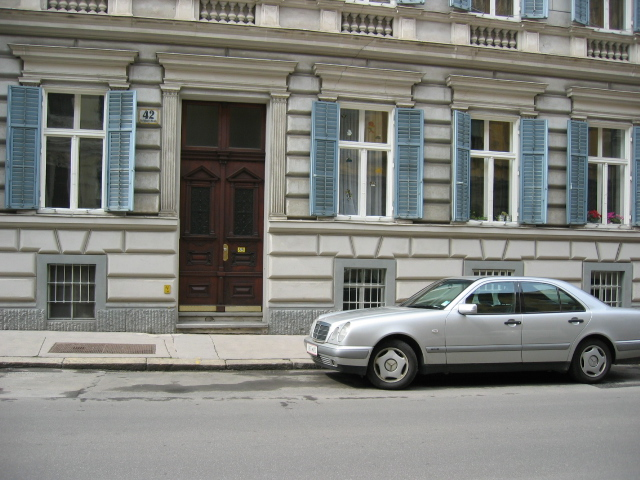}  & \includegraphics[height=2.55cm]{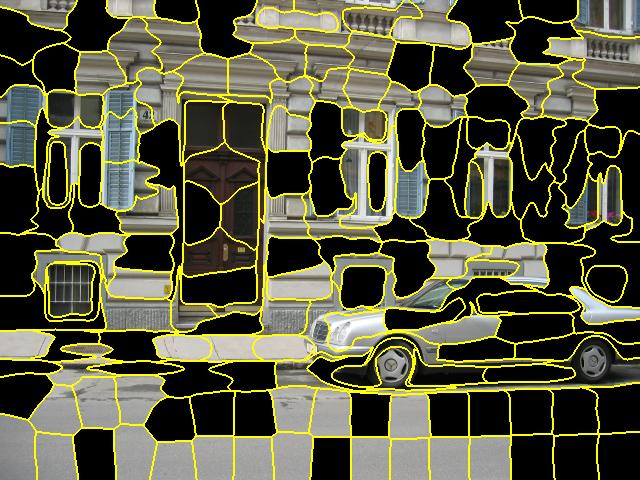} & \includegraphics[height=2.55cm]{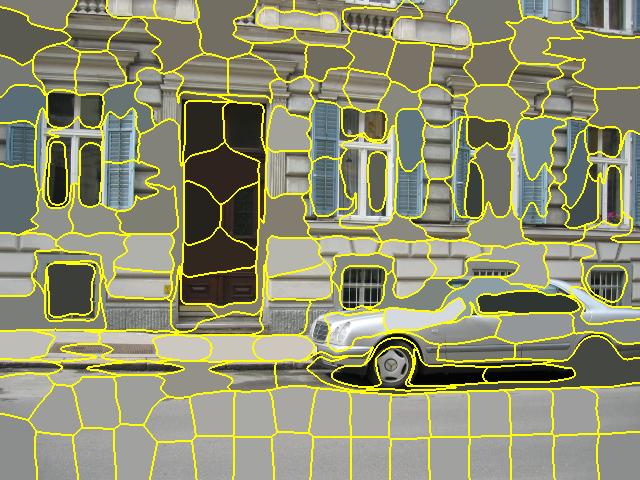}  & \includegraphics[height=2.55cm]{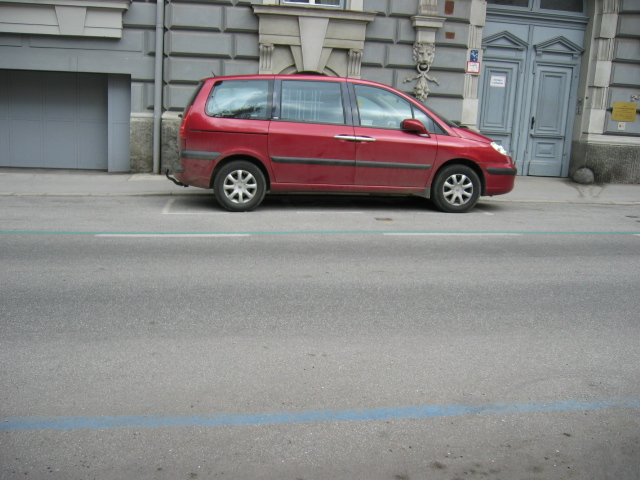}   & \includegraphics[height=2.55cm]{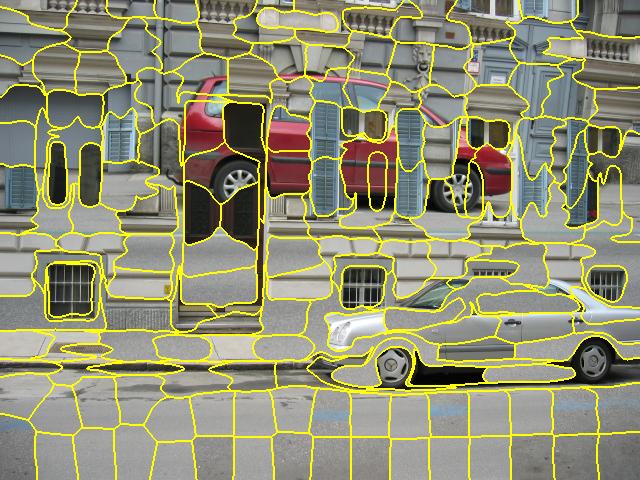}\\
(a) Image 1 & (b) SuperpixelGridCut & (c) SuperpixelGridMean & (d) Image 2 & (e) SuperpixelGridMix\\
\end{tabular}}
\captionof{figure}{(b) and (c) show generated augmented images using the proposed \textit{SuperpixelGridCut} and \textit{SuperpixelGridMean} applied from the original image (a). (e) shows the generated augmented image using the \textit{SuperpixelGridMix} over the original images (a) and (d).\label{teaser}}
}
\makeatother

\maketitle
\thispagestyle{plain}
\pagestyle{plain}



\vspace*{-0.2cm}
\begin{abstract}
\vspace{-0.2cm}
A novel approach of data augmentation based on irregular superpixel decomposition is proposed.
This approach called SuperpixelGridMasks permits to extend original image datasets that are required by training stages of machine learning-related analysis architectures towards increasing their performances. Three variants named  \textit{SuperpixelGridCut}, \textit{SuperpixelGridMean} and \textit{SuperpixelGridMix} are presented. These grid-based methods produce a new style of image transformations using the dropping and fusing of information. Extensive experiments using various image classification models and datasets show that baseline performances can be significantly outperformed using our methods. The comparative study also shows that our methods can overpass the performances of other data augmentations. Experimental results obtained over image recognition datasets of varied natures show the efficiency of these new methods. \textit{SuperpixelGridCut}, \textit{SuperpixelGridMean} and \textit{SuperpixelGridMix} codes are publicly available at {\color{magenta}\href{https://github.com/hammoudiproject/SuperpixelGridMasks}{https://github.com/hammoudiproject/SuperpixelGridMasks}}.
\vskip 0.1cm
\noindent {\small{\normalfont\textbf{Keywords:} classification, YOLO, CNN, SuperpixelGridCut, SuperpixelGridMean, SuperpixelGridMix, deep learning, XAI.}}
\end{abstract}



\begin{table}[t]
\centering\vspace{0.5cm}
{
\resizebox{\columnwidth}{!}{%
\begin{tabular}[t]{ccc}
\toprule 
\textbf{Dataset} & \textbf{Baseline} (\%) & \textbf{Ours} (\%)\\
\midrule
\textit{PASCAL VOC}{\protect\footnotemark} &  57.81 & 75.00 \textbf{(+17.19)}\\
\hline
\textit{Chest X-Ray Images}{\protect\footnotemark} (Kaggle) & 78.68 & 88.14 \textbf{(+9.46)}\\
\bottomrule
\end{tabular}}
}
\caption{Reported top-1 accuracies of our classification results using VGG19 over image recognition datasets of different natures.}
\label{teasertab}
\end{table}%

\vspace*{-1cm}

\section{Introduction}

More than ever, machine learning is a field of great interest for the community of image recognition. To the past, a huge amount of approaches have been developed to perform semi-automatic and automatic image segmentation, image classification, object detection (e.g. YOLO) and so on. Nowadays, deep Convolutional Neural Networks (CNNs) appear as an ultimate resource towards operating recognition tasks over e.g. biomedical images, scene images, object images, document images.

In this context, the investigated topics are more and more refined and delicate to tackle. The CNN architectures often include millions of parameters, requiring an important number of images for performing their training, validation and test stages towards returning decisions with high precisions. In many cases, we face a lack of data due to the topics that are new and that require hand-made and fastidious data labeling tasks. An insufficient amount of relevant data can lead to serious over-fitting problems. Data augmentation appears as a complementary technique for overcoming such problems.  

Various types of data augmentation have been proposed in the literature. Frequent methods can be divided into three categories, i) the pixel-level transforms (e.g. colorimetry-based), ii) the spatial-level transforms (e.g. geometry-based), and iii) the information dropping and fusing (e.g., mask-based). For instance, pixel-level transforms can be obtained by applying conventional image filters such as blur noise, contrast or compression related ones. Spatial-level transforms can be to resize, rotate or flip images. The information dropping and fusing is the most recent category which includes transforms such as image masking or mixing. Each category contains transforms which can integrate parameters used with random values for exploration or optimization reasons. 

The nature of the considered topic, associated images and applicative conditions often help in the choice of the data augmentation methods that have to be experimented. For instance, if the recognition has to be operated through a camera system, then augmenting data by simulating a camera noise via a blur (category 1) could be efficient; if the images to classify represent faces, an horizontal image flip (category 2) could be a good option and a vertical flip not; if the localization of objects in the image of a scene plays an important role, augmenting data by cropping and masking image parts (category 3) can permit to increase the perception field of the considered network by occluding some image parts and forcing then the attention on potentially relevant object parts.

\textbf{Motivation} --- Information dropping techniques (category 3) can present the advantage of forcing a considered network to explore image parts of interest through this trade-off in between image occluding and preserving. Although often efficient and flexible, it is worth mentioning that most of the masking data augmentations methods, i.e. CutOut methods usually exploit the dropping of squares or rectangles.
However, in a real-world topic, most of the processed image sets appear like natural images. The presence of structures that are horizontally or vertically aligned with respect to image orientations rarely occurs. Also, information fusion techniques often aim to merge in one image the image parts of two or more images. It can result from whole image encapsulation or mapping of image parts. The advantage can lie in the generation of new images which embed discriminative characteristics of original image pairs belonging to the same class. Once again, although effective and flexible, most of these mixing data augmentation methods, i.e CutMix methods usually exploit quadrilateral structures. 

In this paper, we propose SuperpixelGridMasks, namely \textit{SuperpixelGridCut}, \textit{SuperpixelGridMean} and \textit{SuperpixelGridMix} data augmentation methods. By comparison to previously presented methods, they have several advantages, namely 
\begin{itemize}
\item to respectively perform dropping and fusing information while limiting the integration in the considered network of periodic linear and right-angles related noises due to the use of regular structures,
\item to generate a 2D irregular structure which is purely built using natural image information (e.g. pixel intensities, textures or contours),
\item to potentially produce structures which delimit real boundaries in between objects and then perform dropping and fusing of delineated object parts instead of dropping and fusing of inconsistent quadrilateral image parts.
\end{itemize} 

\section{Related Work} 

Most of the conventional data augmentation methods and variants from the categories of pixel-level transforms and spatial-level transforms are regrouped in \cite{info11020125}. Data augmentation methods acting on the information dropping and fusing are more and more present. Those previously described such as CutOut, CutMix as well as many variants are presented in \cite{CutOut}, \cite{CutMix} and \cite{SnapMix, intraclassCutMix, MosaicMix, GridMask, GridMix, ChessMix, PuzzleMix, AttentiveCutMix, SaliencyMix, RecursiveMix, CutPaste, AugMix}, respectively. These latter, which are also mainly discussed in this survey \cite{Survey}, are all potentially efficient although they all exploit regular and quadrilateral structures.

To the best of our knowledge, except for direct segmentation purposes, very few approaches exploit superpixel entities towards performing data augmentation tasks. In \cite{info11020125,pmlr-v102-zhang19a}, the authors proposed methods that augment data by transforming input images into superpixel representations which are densely posterized. In \cite{app10248833}, a superpixel representation is produced. Then, each image is split into a set of subimages (i.e. individual superpixels) towards preparing patch-based processes. A method \textit{Superpixel-Mix} is presented in \cite{franchi:hal-03400621}. A superpixel representation is produced and a swap is operated in between some superpixel regions of image pairs. This CutMix is locally operated for augmenting data towards segmentation tasks. 
In sum, current superpixel-based data augmentation either applies raw image posterization or applies localized operations such as image-to-patch decomposition or parsimonious superpixel mixing. 

In our case, superpixel grid-based masks are proposed for the first time to perform global operations by sparsely applying dropping or fusion operations over a superpixel grid with a uniform distribution. This principle is the core of the proposed SuperpixelGridMasks approach which is described in detail in the next section.

\section{Proposed Approach}

The SuperpixelGridMasks approach is presented through three variants; namely \textit{SuperpixelGridCut}, \textit{SuperpixelGridMean} and \textit{SuperpixelGridMix}. Their basis principles are presented in functions \ref{function1}, \ref{function2} and \ref{function3}, respectively. Common initial stage consists of using a superpixel segmentation technique over an original image by providing the number $q$ of requested superpixels. A variety of superpixel segmentation techniques can be used to perform this task (see \cite{WANG201728}). In our case, we selected a refined version which maximizes the matching in between the generated superpixel grid lines and intrinsic contour lines of the original images. By this way, subsequent dropping and fusing processes can operate with coherence on superpixels since they directly share boundaries with object boundaries. The principle consists then of parsing the obtained superpixel list. 

\begin{figure}[t]
\centering
\includegraphics[scale=0.8]{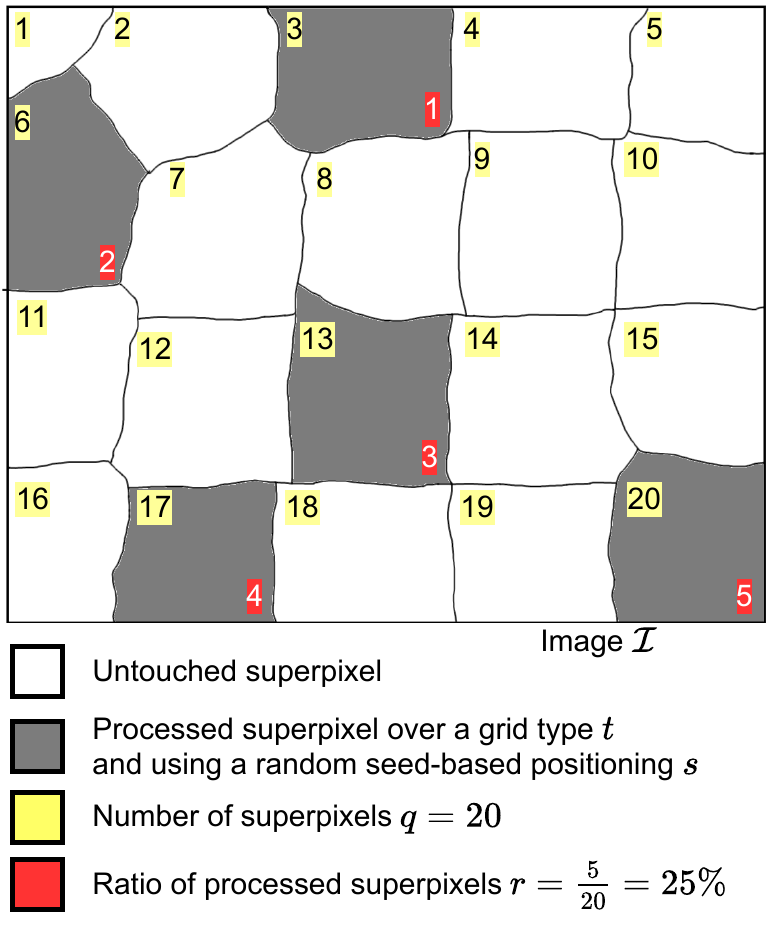}
\caption{Parameters of SupepixelGridMasks approach: segmentation type t , random seed-based positioning s, nb. of superpixels q, ratio of processed superpixels r}
\label{parameters}
\end{figure}

Parameters related to our SuperpixelGridMasks are displayed in Figure \ref{parameters}. A sample of superpixels is then processed (dropped or mixed) depending on a selected occupancy threshold $r$, $r\in \left[0,1\right]$. This rate of processed superpixel occupancy is approximately equal to the rate of pixels to process over the augmented image when superpixel sizes are close. If $r$ is equal to 0.5, then it means that we target the processing of half of superpixels. If $r$ is equal to 0.3, then it means that we target the processing of 30\% of the list of superpixels. To perform a uniform superpixel processing, we proceed by randomly drawing a variable $p_{i}$ from a uniform distribution over $\left[0,1\right]$ for each superpixel $i$. The processing decision $d$ is finally given by Eq.(\ref{E0}). 

\begin{equation}
\label{E0}
d =  \left\{\begin{array}{l}
processed~superpixel~if~p_{i} \in \left[0,r\right]\\ 
unprocessed~otherwise\\
\end{array}\right.
\end{equation}

if $p$ is inferior to the selected $r$, then the considered superpixel will be processed. Otherwise, the superpixel will be untouched. This random pulling of $p$ permits to process the superpixel in a sparse manner and with a uniform distribution. By this way, the threshold $r$ also permits to control the contiguity level of processed superpixels.

    \begin{algorithm}[t]
		\SetAlgorithmName{Function}{}
        \SetAlgoLined
        \SetKwInOut{Input}{Input}
        \SetKwInOut{Output}{Output}
        \Input{I:\ original~image; q: nb of Superpixels\; r: rate of processed superpixel occupancy}
        \Output{J: generated~SuperpixelGridCut~image}  
        \BlankLine
				J$\leftarrow$I\;
				\tcc{Superpixel segmentation}
				List Superpixel $\leftarrow$ Superpixel\_segmentation(J,q)\;
				\tcc{Superpixel parsing \& cutting}
				\For {($i=1; i\leq Superpixel.size(); i++$)}
				{
				$p=rand(0,1)$\;
				\If {($p<r$)}
				{Superpixel(i) $\leftarrow$ Black}
				}
        \Return $J$
    \caption{SuperpixelGridCut(I,q,r)}
	\label{function1}
    \end{algorithm}

  \begin{algorithm}[t]
		\SetAlgorithmName{Function}{}
        \SetAlgoLined
        \SetKwInOut{Input}{Input}
        \SetKwInOut{Output}{Output}
        \Input{I:\ original~image; q: nb of Superpixels\; r: rate of processed superpixel occupancy}
        \Output{J: generated~SuperpixelGridMean~image}  
        \BlankLine
				J$\leftarrow$I\;
				Integer AVG\;
				\tcc{Superpixel segmentation}
				List Superpixel $\leftarrow$ Superpixel\_segmentation(J,q)\;
				\tcc{Superpixel parsing \& averaging}
				\For {($i=1; i\leq Superpixel.size(); i++$)}
				{
				$p=rand(0,1)$\;
				\If {($p<r$)}
					{
					AVG $\leftarrow$ mean(Superpixel(i))\;
					Superpixel(i) $\leftarrow$ AVG\;
					}
				}
        \Return $J$
    \caption{SuperpixelGridMean(I,q,r)}
	\label{function2}
    \end{algorithm}

\begin{algorithm}[t]
		\SetAlgorithmName{Function}{}
        \SetAlgoLined
        \SetKwInOut{Input}{Input}
        \SetKwInOut{Output}{Output}
        \Input{I,J:\ original~images; q: nb of Superpixels\; r: rate of processed superpixel occupancy}
        \Output{K: generated~SuperpixelGridMix~image}  
        \BlankLine
				K$\leftarrow$I\;
				\tcc{Superpixel segmentation of K}
				List Superpixel $\leftarrow$ Superpixel\_segmentation(K,q)\;
				\tcc{Superpixel parsing}
				\For {($i=1; i\leq Superpixel.size(); i++$)}
				{
				$p=rand(0,1)$\;
				\If {($p<r$)}
					{
					\tcc{Pixel parsing \& copying}
					\For {($j=1; j\leq Superpixel(i).size(); j++$)}
						{
						Pixel$_{K}(j)$ $\leftarrow$ Pixel$_{J}(j)$
						}
					}
				}
        \Return $K$
    \caption{SuperpixelGridMix(I,J,q,r)}
	\label{function3}
    \end{algorithm}

From this step, function \ref{function1} and function \ref{function2} directly colorize the retained sample of superpixels to process in black or with average color values, respectively while the function \ref{function3} replaces the texture of the retained sample of superpixels to process by the one of another original image of the dataset. In this latter process, the basis version employs a second image that belongs to the same image class than the first image and a correspondence in between pixel positions is preserved during the texture mapping. 

The methods \textit{SuperpixelGridCut}, \textit{SuperpixelGridMean} and \textit{SuperpixelGridMix} can be unified using the expression:

\begin{equation}
\label{E1}
K = (1-M) \odot I + M \odot J 
\end{equation}

\noindent where $\odot$ is the element-wise operation, $A = (Superpixel_A(1), \cdots, Superpixel_A(q))^t$ $A \in \{I, J, K\}$.
$M = (M(1), \cdots, M(q))^t$  with 
\begin{equation}
\label{E2}
M(i) =  \left\{\begin{array}{l}
0~if~Superpixel(i)~untouched\\ 
1~otherwise\\
\end{array}\right.
\end{equation}

\noindent In Eq.(\ref{E1}), function \ref{function1} uses $J = 0$ and function \ref{function2} employs the image $J$ defined by:

\begin{equation}
\label{E3}
Superpixel_J(i) = mean(Superpixel_I(i))
\end{equation} 

Finally, in case of function \ref{function3}, the image $J$ is an image different than $I$, selected from the dataset.

Besides, by applying the \textit{SuperpixelGridCut} (or \textit{SuperpixelGridMean}), a one-to-one augmentation is obtained in the sense that each original image can produce one augmented image. By this way, the size of a considered training image set can then be multiplied by $2$. By applying the \textit{SuperpixelGridMix} to a training set of $N$ images, each image can potentially be mixed with $N-1$ images. The number of images in the training set after augmentation can reach a number: 
\begin{equation}
\label{E4}a = N + N \times (N-1)= x^{2}
\end{equation}
where $N^{2}-N$  images are augmented ones. We remind that in this latter case, each original image of the training set has its own generated superpixel grid. These calculations assume the use of our augmentation methods in their conventional forms. 

Further augmented images can be obtained by varying parameters shown in Figure \ref{parameters} such as instances of the random positioning for processed superpixels (e.g. using other seeds $s$), the superpixel quantity $q$, the threshold of processed superpixel occupancy $r$ or even the superpixel grid type $t$ by applying other superpixel segmentation methods.

\section{Experimental Results and Evaluation}

\subsection{Datasets, Implementation Details and Protocol}

For our experimental needs, we exploited two image datasets of different natures that are publicly available, namely a \textit{PASCAL VOC} dataset and the \textit{Chest X-Ray Images} dataset. PASCAL VOC{\protect\footnotemark[1]} corresponds to a standardized image set for object class recognition e.g. towards urban scene analysis. For simulating conditions for which a classification problem is new and has a limited number of labeled images, we used the PASCAL VOC{\protect\footnotemark[1]} dataset which is composed of four image classes; namely, motorbikes, bicycles, people, and cars in arbitrary pose. A subset of 679 color images having varied resolutions has been experimented. Besides, \textit{Chest X-Ray Images}{\protect\footnotemark[2]} is a dataset which is exploited in biomedical studies for pneumonia analysis. This latter includes two specific image classes; namely, normal and viral or bacterial pneumonia. Chest X-Ray Image contains 5856 gray-scale images having relatively high resolutions. It is worth mentioning that the visual classification of such a dataset is extremely difficult for a non-expert since one single image type is represented, namely chest X-Ray images. Hence, discussed augmentation methods are experimented over two challenging classification tasks.

A VGG19 classification architecture has been exploited since this latter is often used as a comparison baseline in image classification studies. In our case, the objective is not to outperform state-of-the-art results by comparing various CNNs performances over a specific classification problem, but observing the basis performance over a reference classification architecture and the impact of data augmentation methods over varied datasets. 

By using the selected PASCAL VOC and Chest X-Ray Images datasets; \textit{SuperpixelGridCut}, \textit{SuperpixelGridMean} and \textit{SuperpixelGridMix} have then been experimented over binary and multi-class image classification problems. The dataset baselines have been calculated using VGG19 without data augmentation. For measuring the impact of our methods, the training stage has been done from scratch when using the selected PASCAL VOC since some of its object classes (e.g. vehicles) are already well represented in ImageNet \cite{ImageNet}. However, our VGG19 architecture uses an ImageNet pretraining for classifying the Chest X-Ray Images dataset because of its high inter-class similarity. 

\textit{SuperpixelGridCut}, \textit{SuperpixelGridMean} and \textit{SuperpixelGridMix} have employed the SLIC superpixel segmentation method \cite{Fua} for generating the superpixel grids with close superpixel sizes. The data augmentations \textit{SuperpixelGridCut} and \textit{SuperpixelGridMean} have been applied to the original images of our training sets permitting them to multiply their sizes by 2. Regarding the \textit{SuperpixelGridMix} data augmentation, it is worth mentioning that it can directly be applied in between original images if they share the same size but a strategy has to be adopted if the image size differs from one image to the other. In our implementation, a segmented original image (i.e. having a generated superpixel grid) uses a second original image for a \textit{SuperpixelGridMix} if this latter has a size that is superior or equal to the segmented one. By this way, border effects are avoided during the processing. Also, we apply this method on the set of consecutive image pairs from the training sets. In this case, the number of augmented images is approximately equal to the size of the original training sets, respectively. 

\begin{figure}[t]
\centering
\includegraphics[scale=0.9]{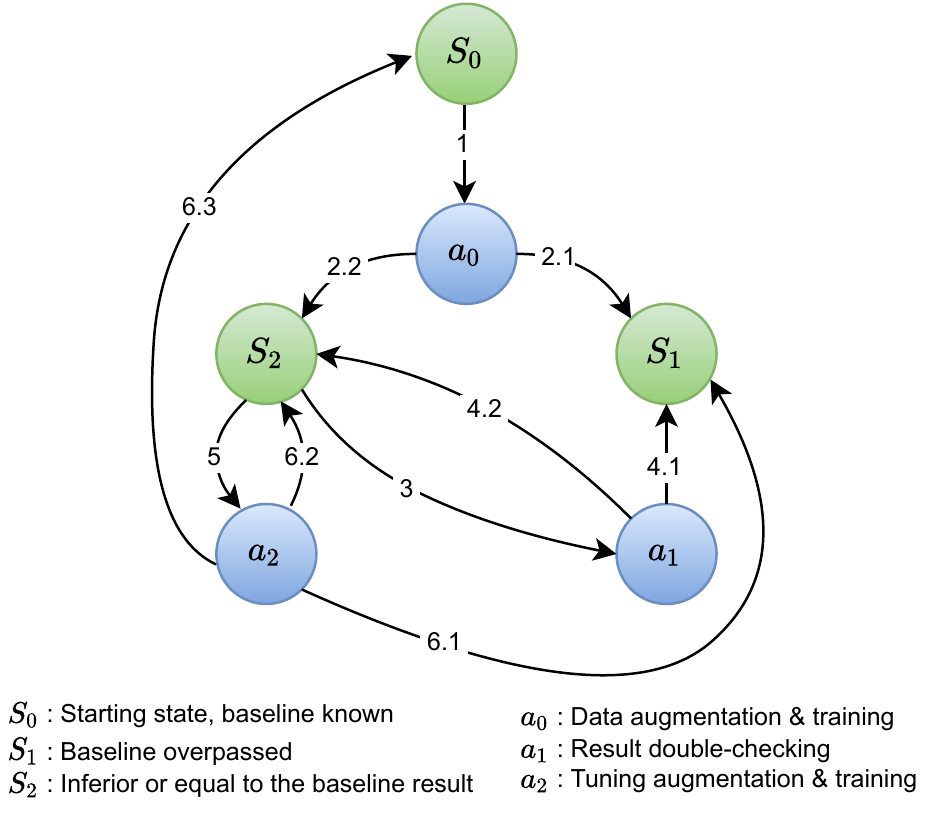}
\caption{State diagram which models an instance of our protocol for classification performance searching. Considered experimental states and actions are represented.}
\label{Statediagram}
\end{figure}

Figure \ref{Statediagram} displays a one-pass scenario of our protocol for classification performance searching. By this way, experimented states and actions are concisely presented. Let us consider a starting state $\mathcal{S}_{0}$ for which the baseline classification result obtained from a query dataset and a selected model are known. First action $a_{0}$ consists of applying a query data augmentation method and training a model. This action can conduct to overpass the baseline result (state $\mathcal{S}_{1}$); otherwise, the result is inferior or equal to the baseline (state $\mathcal{S}_{2}$). Next action $a_{1}$ aims to double-check this latter result. This can conduct to overpass the baseline result (state $\mathcal{S}_{1}$); otherwise, if the result is still inferior to the baseline (state $\mathcal{S}_{2}$), then action $a_{2}$ consists of tuning the augmentation method and training. This can lead to obtain a result better than the baseline (state $\mathcal{S}_{1}$); otherwise the process is restarted (state $\mathcal{S}_{0}$) e.g. acting by using another data augmentation method. 

\begin{figure*}[h!] \centering
\begin{subfigure}{.18\textwidth}
  \centering
  \includegraphics[width=3cm]{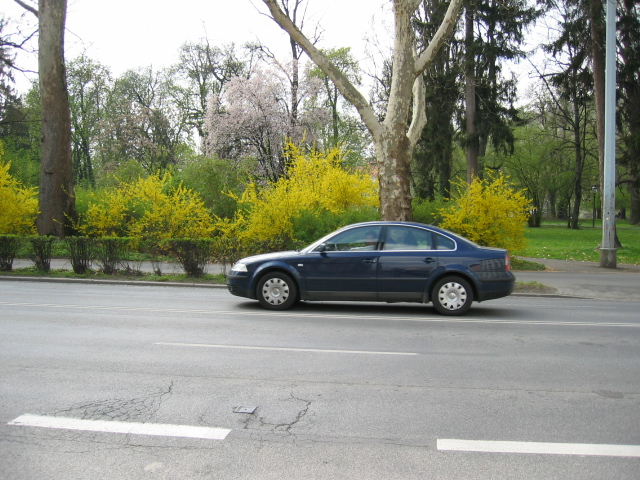}  
  \caption{Image 1.}
  \label{332}
\end{subfigure}
\begin{subfigure}{.18\textwidth}
  \centering
  \includegraphics[width=3cm]{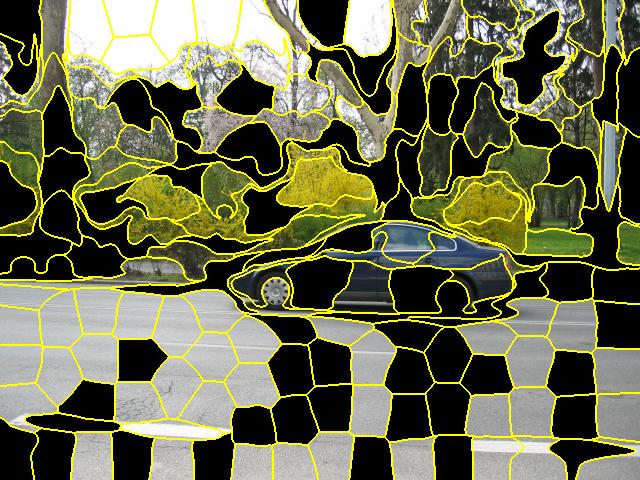}  
  \caption{SuperpixelGridCut.}
  \label{44}
\end{subfigure}
\begin{subfigure}{.18\textwidth}
  \centering
  \includegraphics[width=3cm]{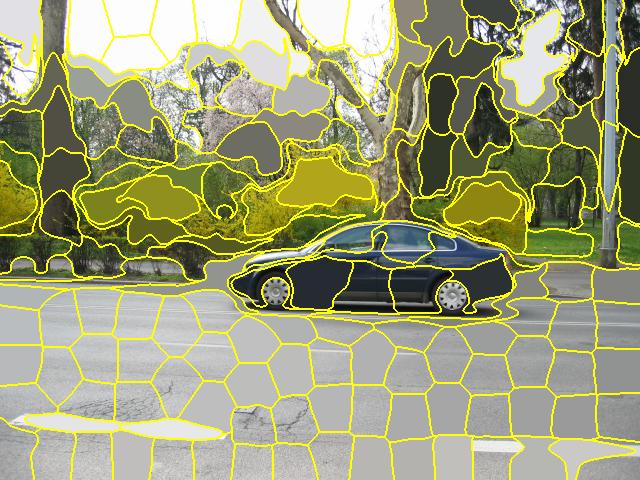}  
  \caption{SuperpixelGridMean.}
  \label{55}
\end{subfigure}
\begin{subfigure}{.18\textwidth}
  \centering
  \includegraphics[width=3cm]{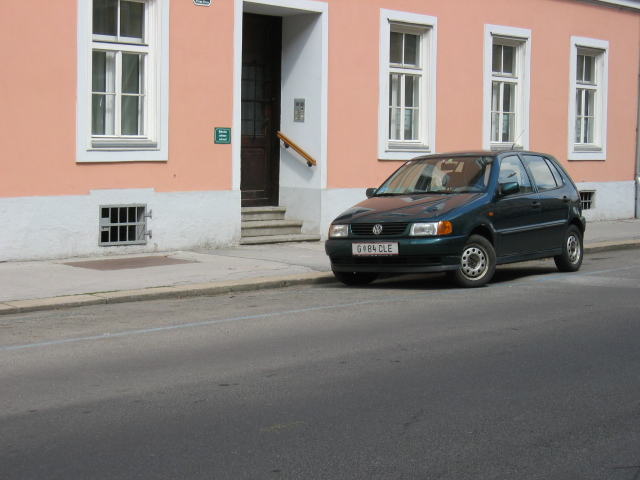}  
  \caption{Image 2.}
  \label{551}
\end{subfigure}
\begin{subfigure}{.18\textwidth}
  \centering
  \includegraphics[width=3cm]{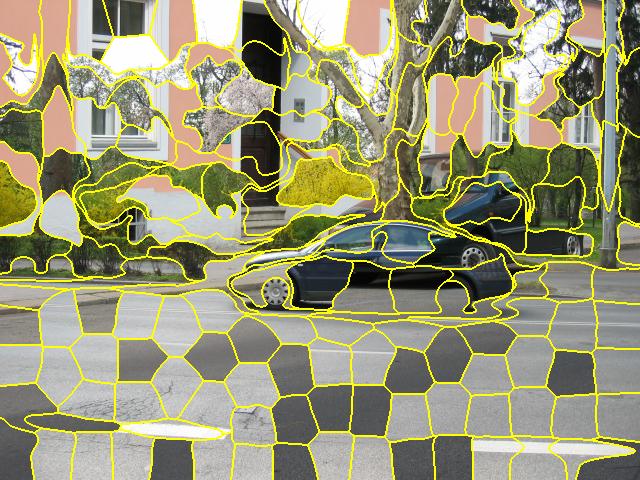}  
  \caption{SuperpixelGridMix.}
  \label{551}
\end{subfigure}\\
\begin{subfigure}{.18\textwidth}
  \centering
  \includegraphics[width=3cm]{carsgraz_227.png}  
  \caption{Image 1.}
  \label{332}
\end{subfigure}
\begin{subfigure}{.18\textwidth}
  \centering
  \includegraphics[width=3cm]{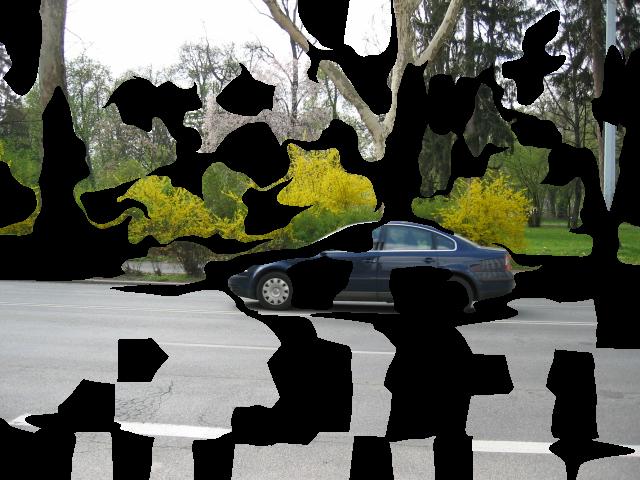}  
  \caption{SuperpixelGridCut.}
  \label{44}
\end{subfigure}
\begin{subfigure}{.18\textwidth}
  \centering
  \includegraphics[width=3cm]{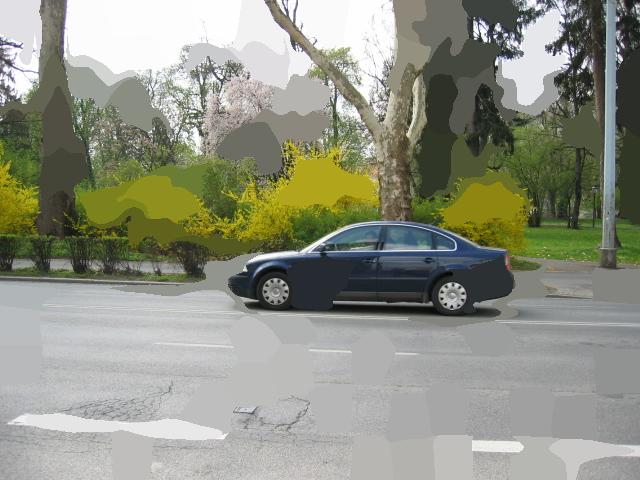}  
  \caption{SuperpixelGridMean.}
  \label{55}
\end{subfigure}
\begin{subfigure}{.18\textwidth}
  \centering
  \includegraphics[width=3cm]{carsgraz_236.png}  
  \caption{Image 2.}
  \label{551}
\end{subfigure}
\begin{subfigure}{.18\textwidth}
  \centering
  \includegraphics[width=3cm]{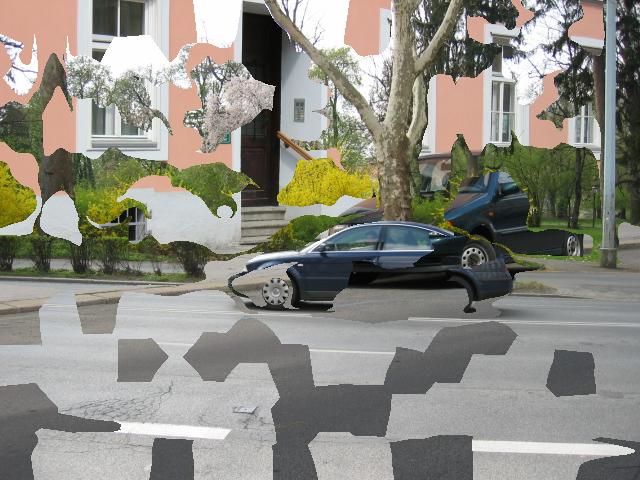}  
  \caption{SuperpixelGridMix.}
  \label{551}
\end{subfigure}
\vspace{0.2cm}
\hrule
\vspace{0.2cm}
\begin{subfigure}{.18\textwidth}
  \centering
  \includegraphics[width=3cm]{carsgraz_227.png}  
  \caption{Image 1.}
  \label{332}
\end{subfigure}
\begin{subfigure}{.18\textwidth}
  \centering
  \includegraphics[width=3cm]{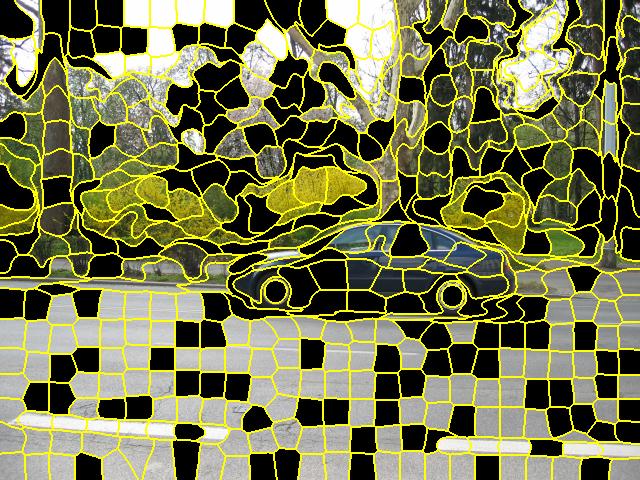}  
  \caption{SuperpixelGridCut.}
  \label{44}
\end{subfigure}
\begin{subfigure}{.18\textwidth}
  \centering
  \includegraphics[width=3cm]{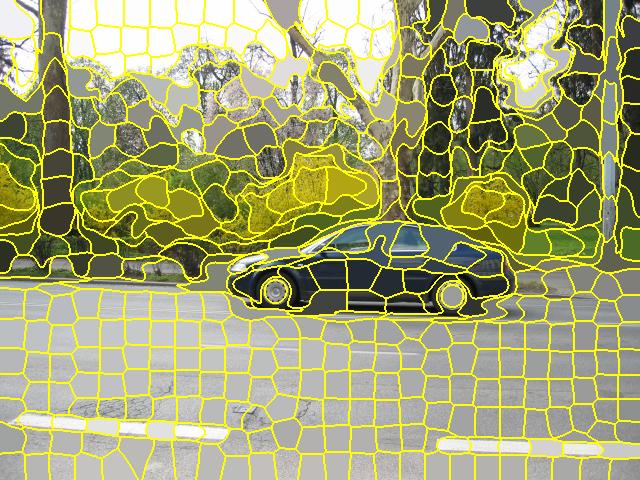}  
  \caption{SuperpixelGridMean.}
  \label{55}
\end{subfigure}
\begin{subfigure}{.18\textwidth}
  \centering
  \includegraphics[width=3cm]{carsgraz_236.png}  
  \caption{Image 2.}
  \label{551}
\end{subfigure}
\begin{subfigure}{.18\textwidth}
  \centering
  \includegraphics[width=3cm]{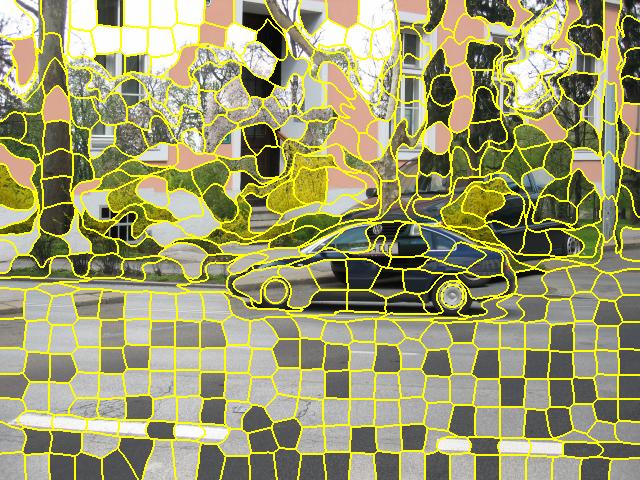}  
  \caption{SuperpixelGridMix.}
  \label{551}
\end{subfigure}\\
\begin{subfigure}{.18\textwidth}
  \centering
  \includegraphics[width=3cm]{carsgraz_227.png}  
  \caption{Image 1.}
  \label{332}
\end{subfigure}
\begin{subfigure}{.18\textwidth}
  \centering
  \includegraphics[width=3cm]{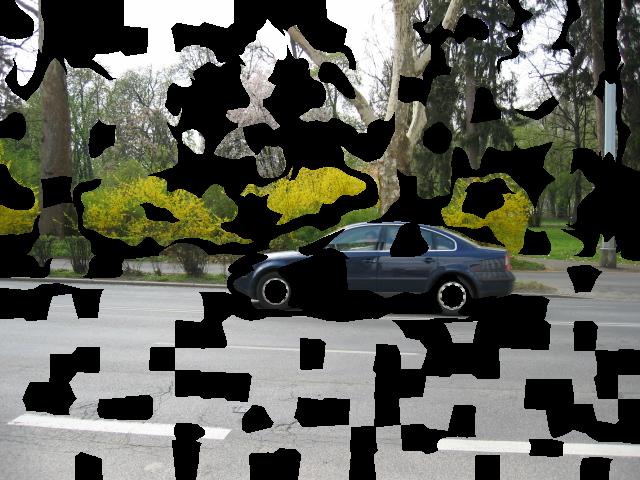}  
  \caption{SuperpixelGridCut.}
  \label{44}
\end{subfigure}
\begin{subfigure}{.18\textwidth}
  \centering
  \includegraphics[width=3cm]{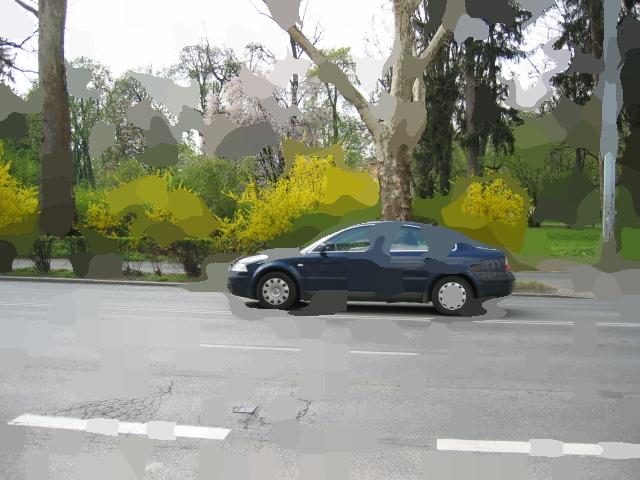}  
  \caption{SuperpixelGridMean.}
  \label{55}
\end{subfigure}
\begin{subfigure}{.18\textwidth}
  \centering
  \includegraphics[width=3cm]{carsgraz_236.png}  
  \caption{Image 2.}
  \label{551}
\end{subfigure}
\begin{subfigure}{.18\textwidth}
  \centering
  \includegraphics[width=3cm]{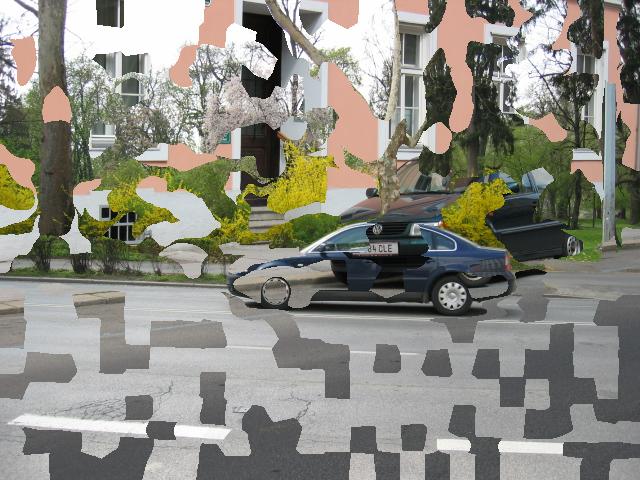}  
  \caption{SuperpixelGridMix.}
  \label{551}
\end{subfigure}
\caption{Augmentations generated by applying the proposed \textit{SuperpixelGrixCut}, \textit{SuperpixelGridMean} and \textit{SuperpixelGridMix} over urban and semi-urban scene images. Top rows and bottom rows show our results with two selected superpixel quantities, $q=200$ and $q=500$, respectively. The selected ratio of processed superpixels is $r=0.4$.}
\label{experiment1}
\end{figure*}

\begin{figure*}[h!] \centering
\begin{subfigure}{.18\textwidth}
  \centering
  \includegraphics[width=3cm]{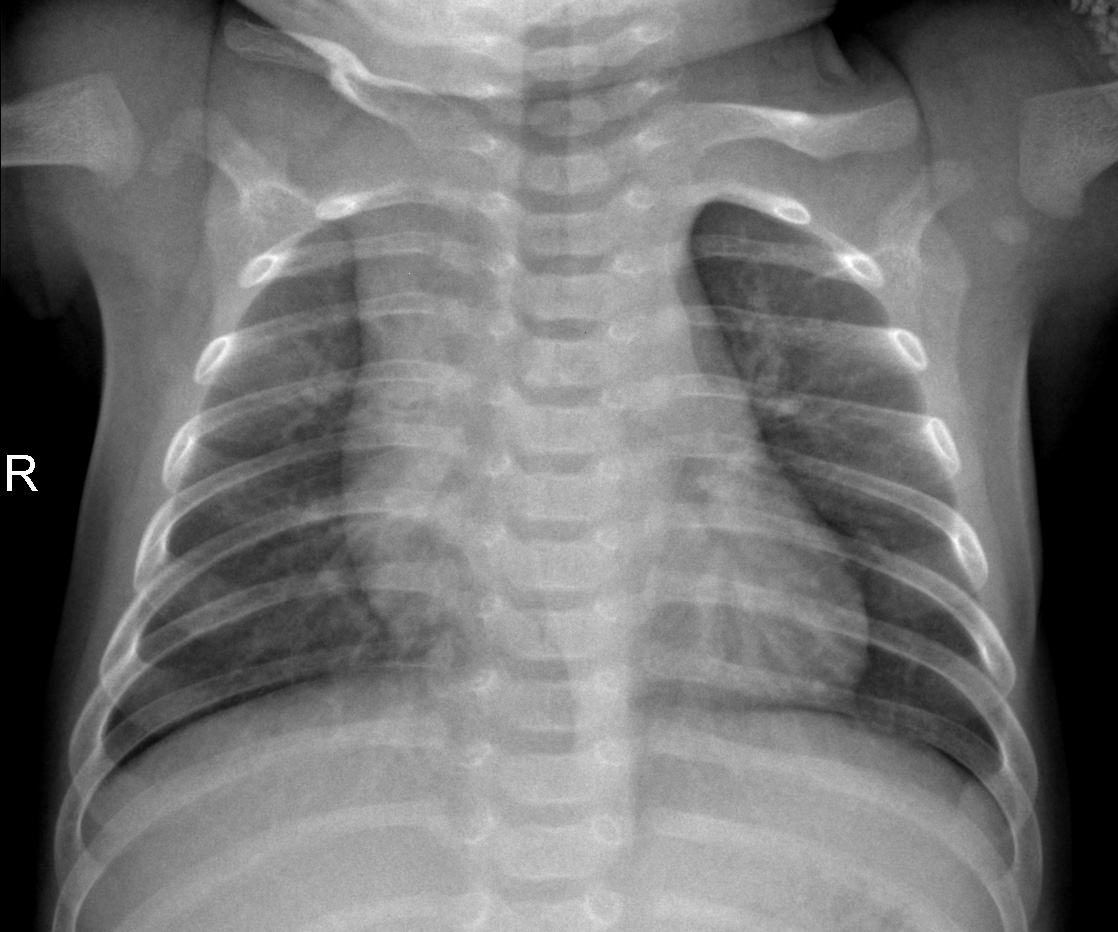}  
  \caption{Image 1.}
  \label{332}
\end{subfigure}
\begin{subfigure}{.18\textwidth}
  \centering
  \includegraphics[width=3cm]{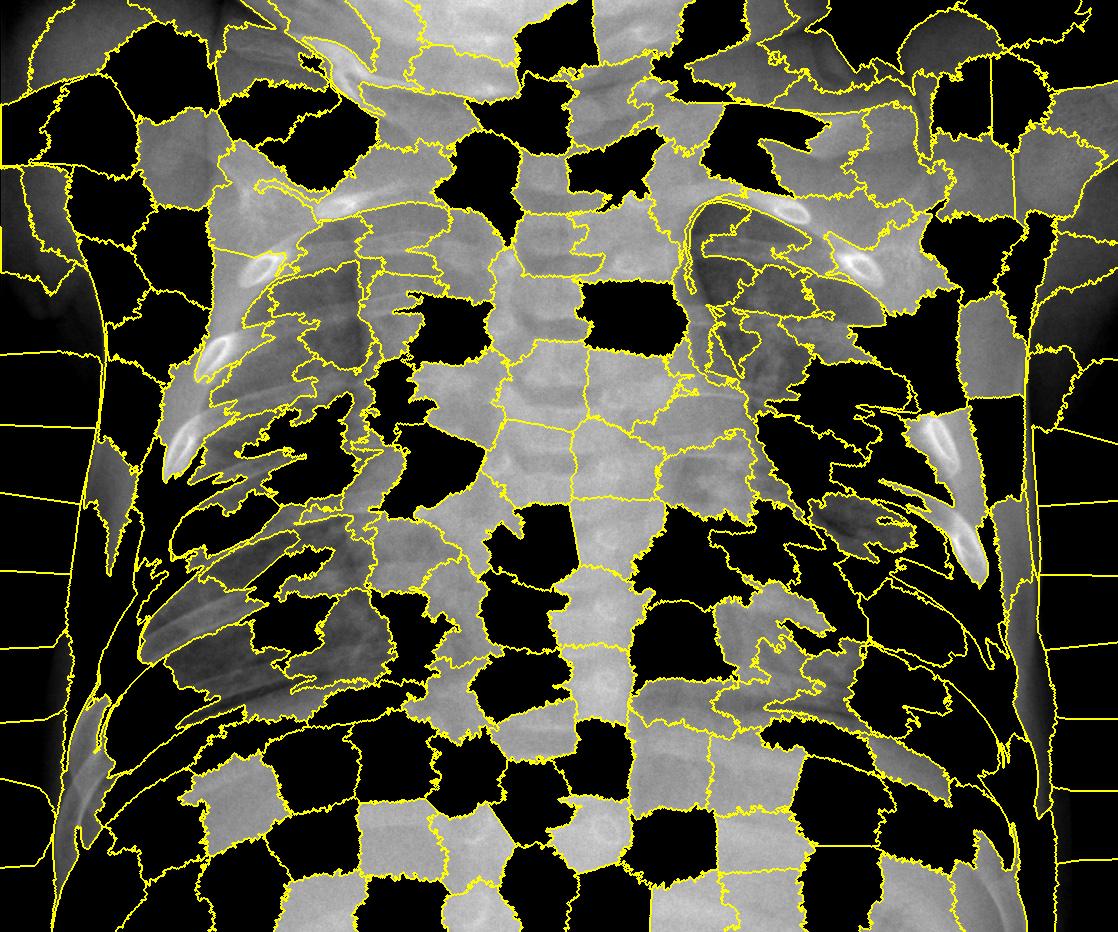}  
  \caption{SuperpixelGridCut.}
  \label{44}
\end{subfigure}
\begin{subfigure}{.18\textwidth}
  \centering
  \includegraphics[width=3cm]{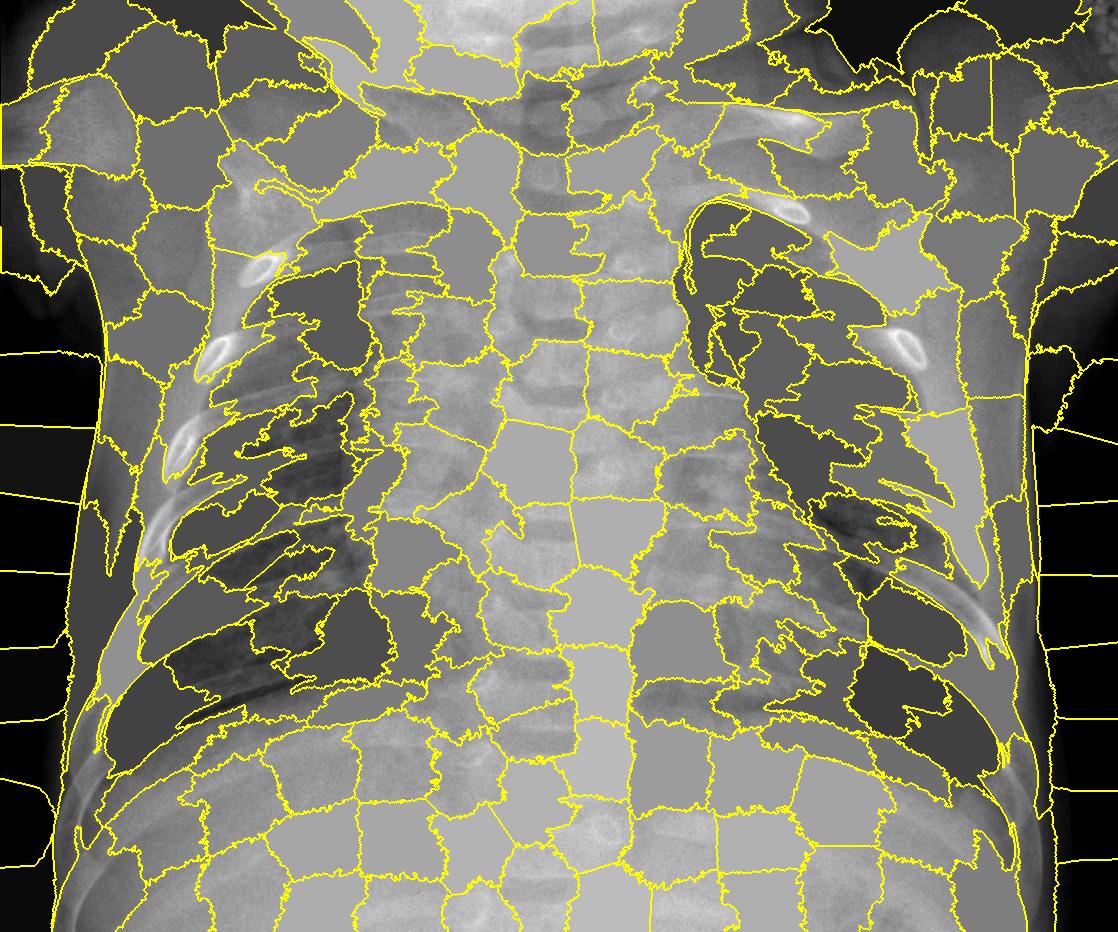}  
  \caption{SuperpixelGridMean.}
  \label{55}
\end{subfigure}
\begin{subfigure}{.18\textwidth}
  \centering
  \includegraphics[width=3cm]{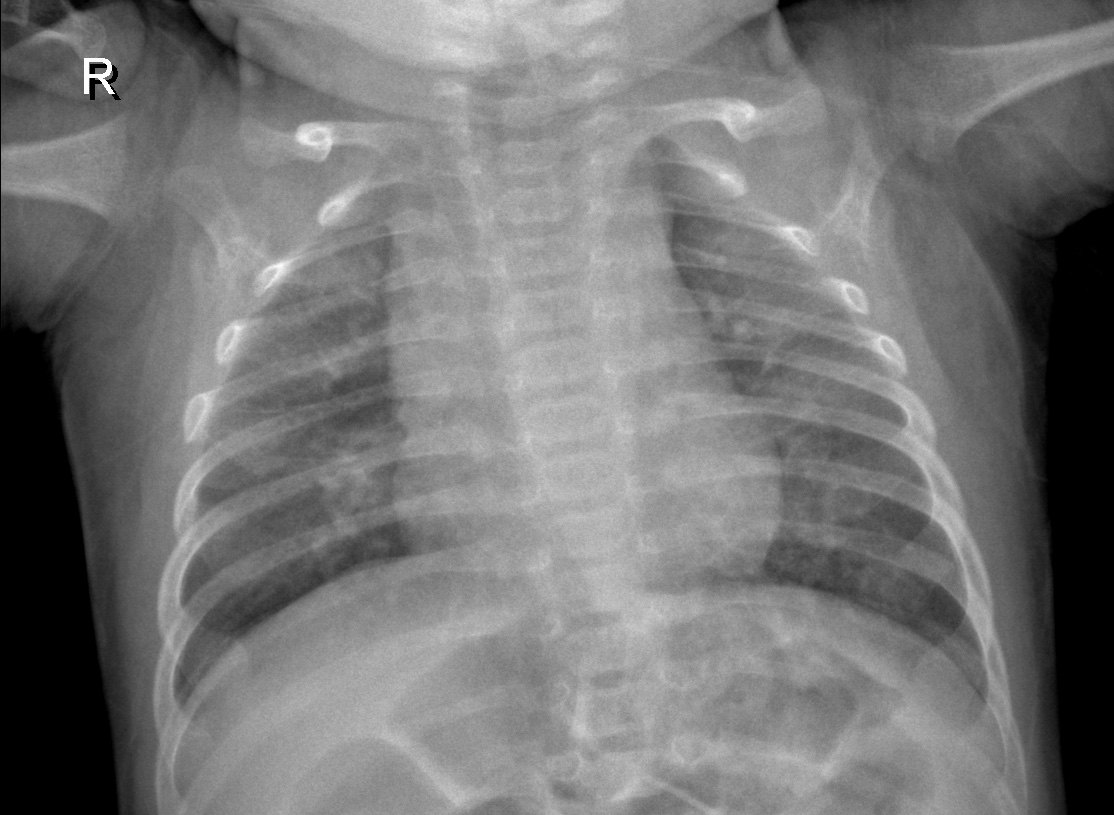}  
  \caption{Image 2.}
  \label{551}
\end{subfigure}
\begin{subfigure}{.18\textwidth}
  \centering
  \includegraphics[width=3cm]{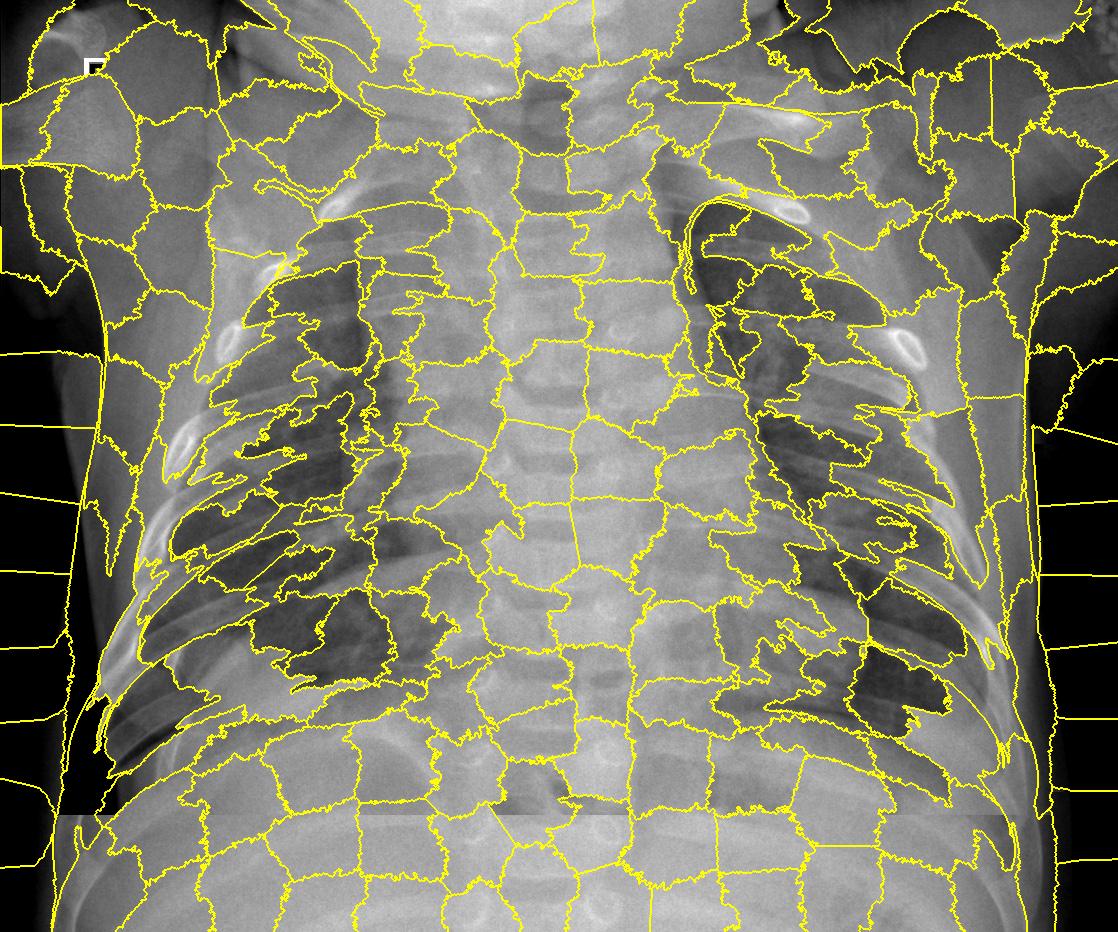}  
  \caption{SuperpixelGridMix.}
  \label{551}
\end{subfigure}\\
\begin{subfigure}{.18\textwidth}
  \centering
  \includegraphics[width=3cm]{IM06930001.jpeg}  
  \caption{Image 1.}
  \label{332}
\end{subfigure}
\begin{subfigure}{.18\textwidth}
  \centering
  \includegraphics[width=3cm]{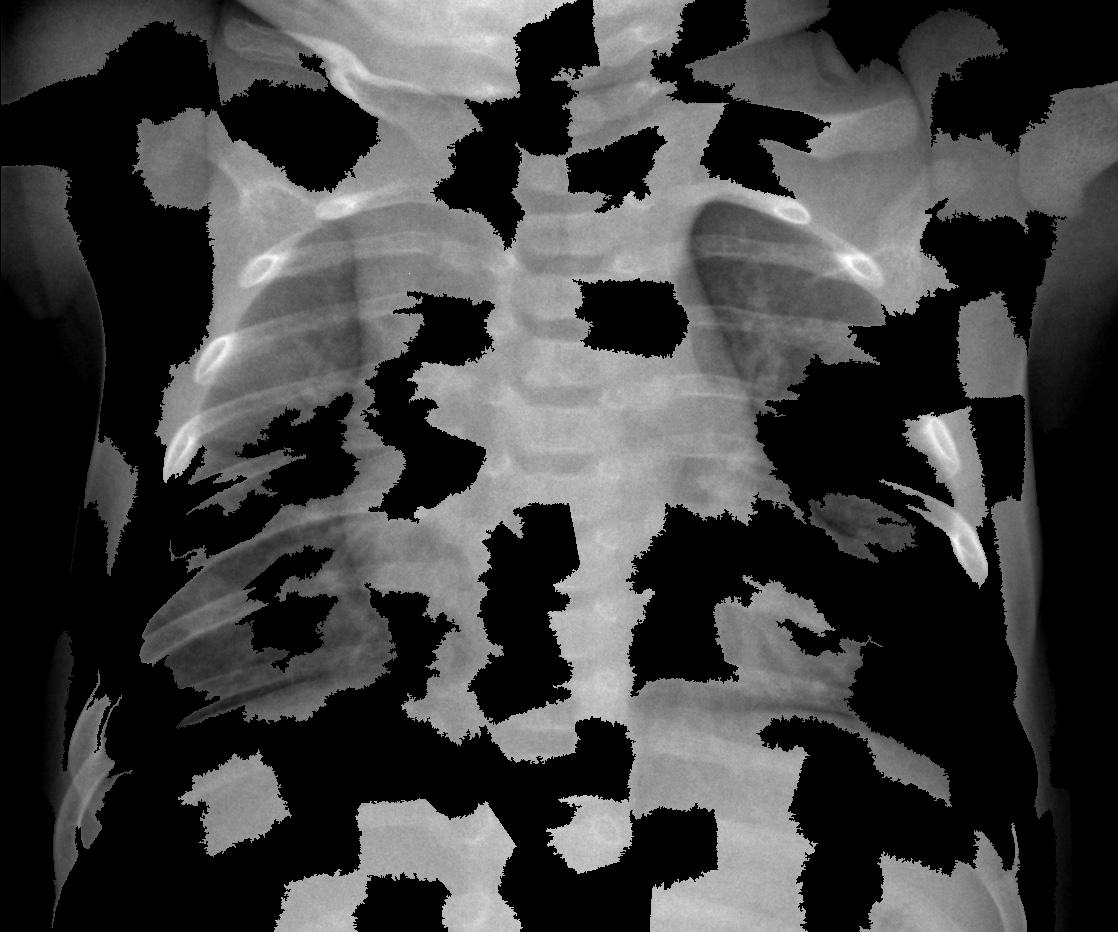}  
  \caption{SuperpixelGridCut.}
  \label{44}
\end{subfigure}
\begin{subfigure}{.18\textwidth}
  \centering
  \includegraphics[width=3cm]{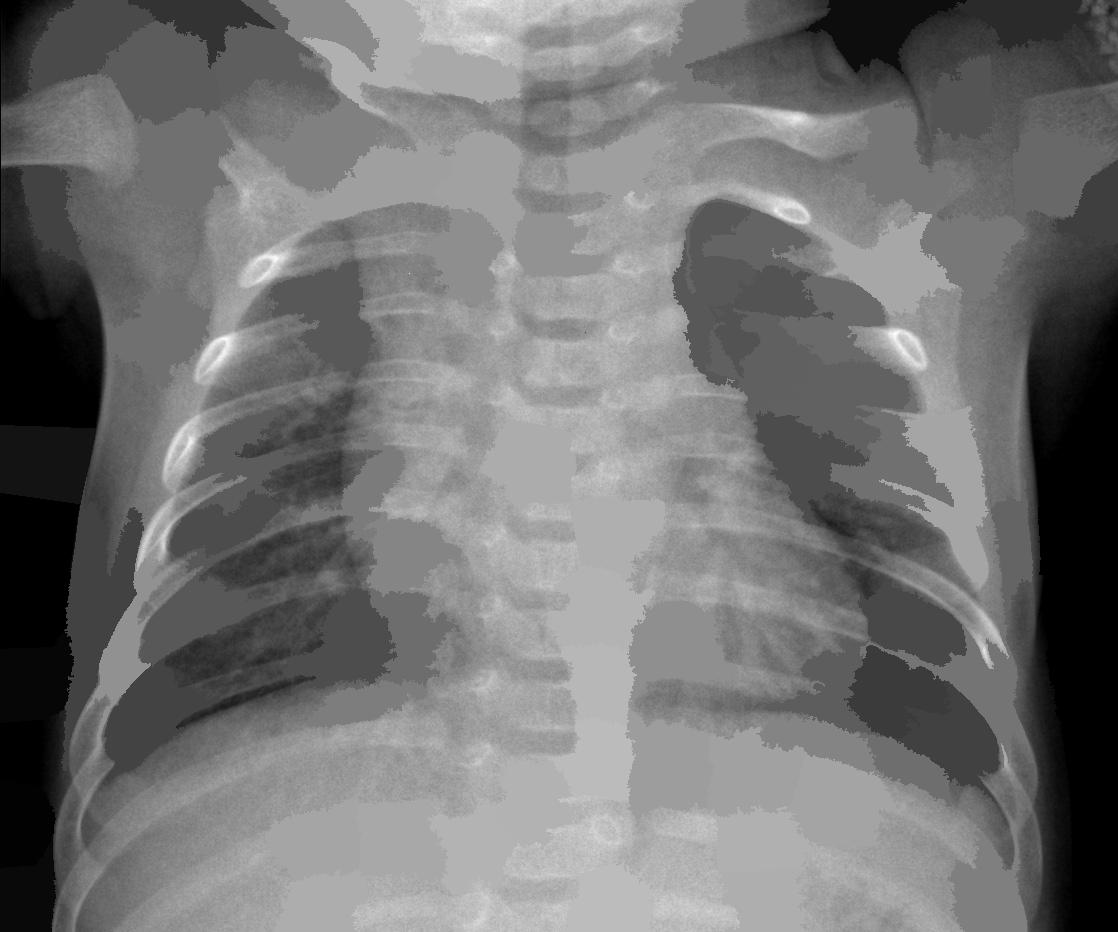}  
  \caption{SuperpixelGridMean.}
  \label{55}
\end{subfigure}
\begin{subfigure}{.18\textwidth}
  \centering
  \includegraphics[width=3cm]{IM07390001.jpeg}  
  \caption{Image 2.}
  \label{551}
\end{subfigure}
\begin{subfigure}{.18\textwidth}
  \centering
  \includegraphics[width=3cm]{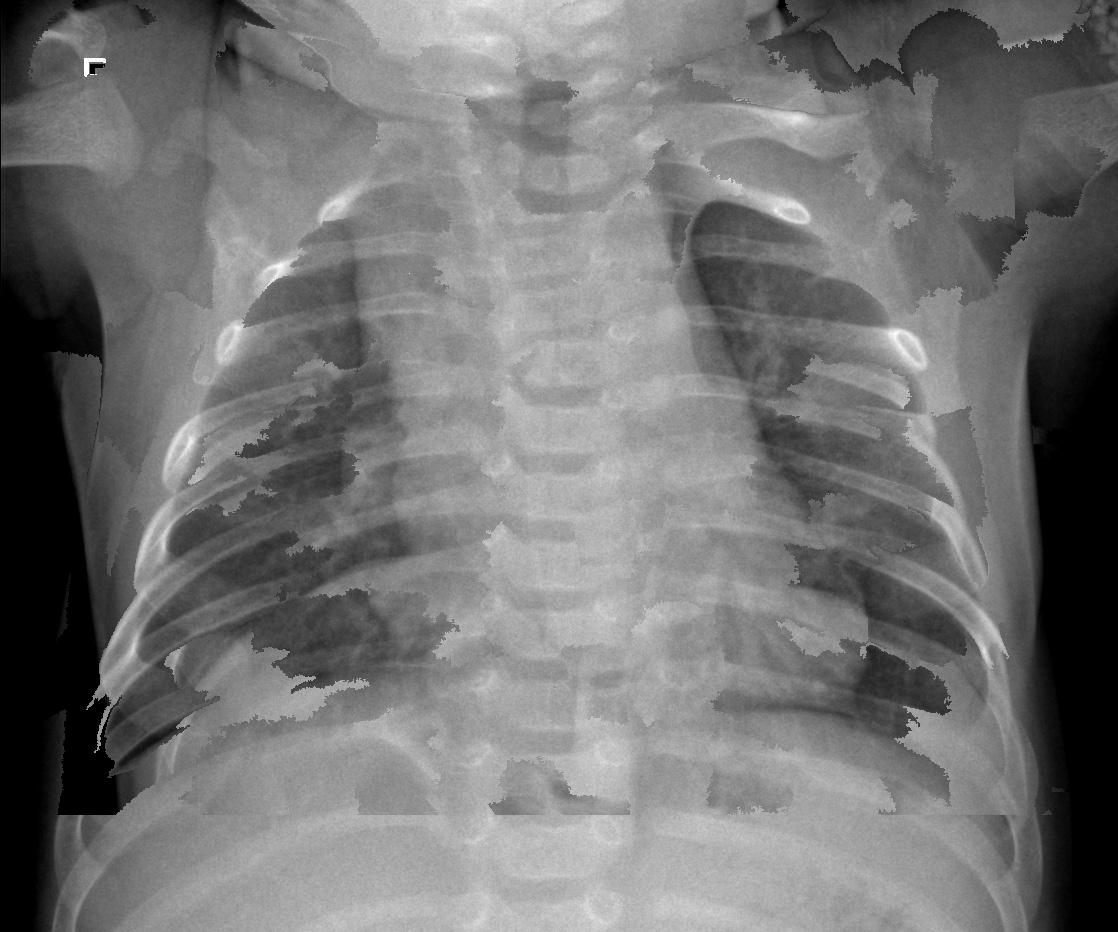}  
  \caption{SuperpixelGridMix.}
  \label{551}
\end{subfigure}
\vspace{0.2cm}
\hrule
\vspace{0.2cm}
\begin{subfigure}{.18\textwidth}
  \centering
  \includegraphics[width=3cm]{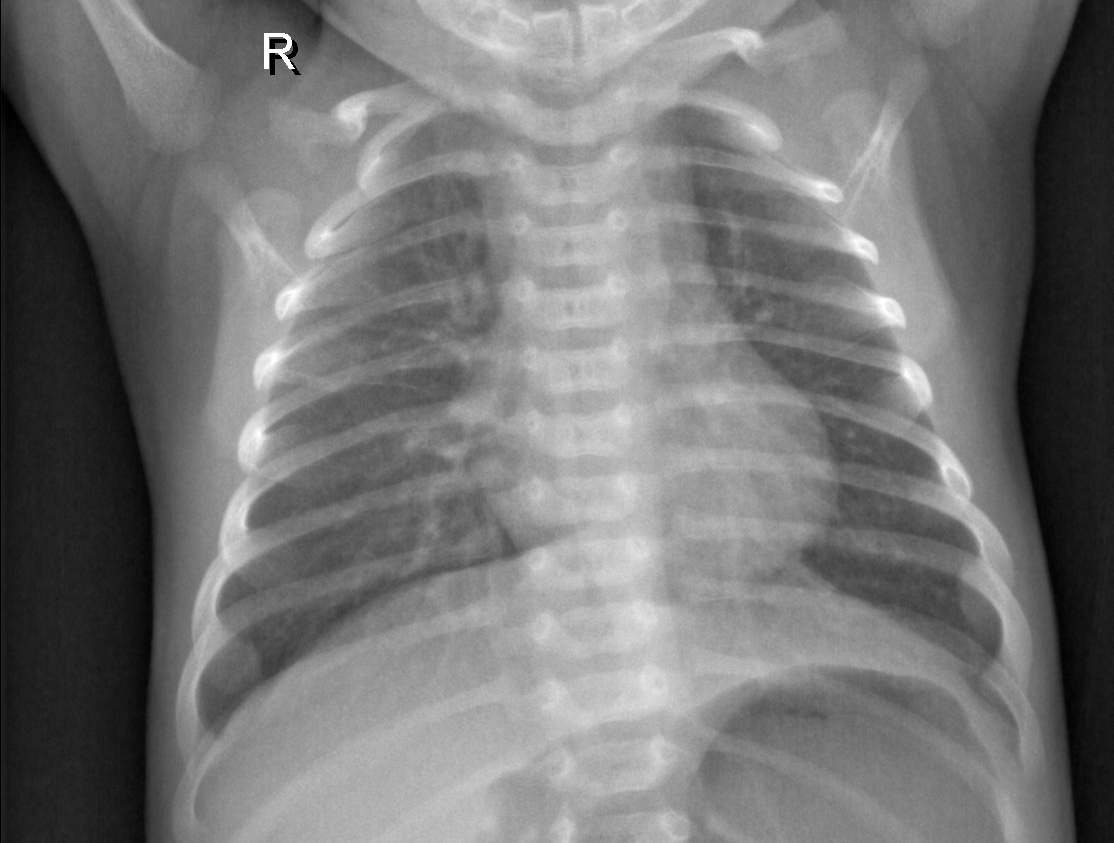}  
  \caption{Image $1^\prime$.}
  \label{332}
\end{subfigure}
\begin{subfigure}{.18\textwidth}
  \centering
  \includegraphics[width=3cm]{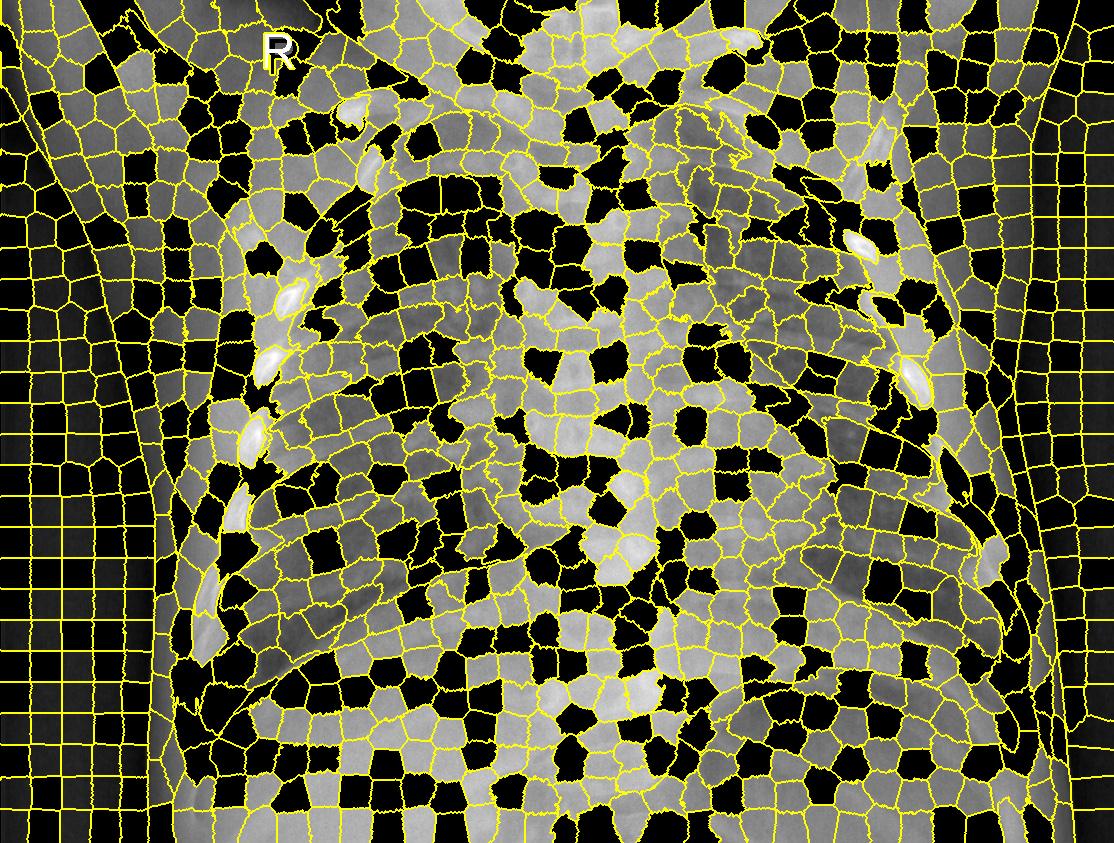}  
  \caption{SuperpixelGridCut.}
  \label{44}
\end{subfigure}
\begin{subfigure}{.18\textwidth}
  \centering
  \includegraphics[width=3cm]{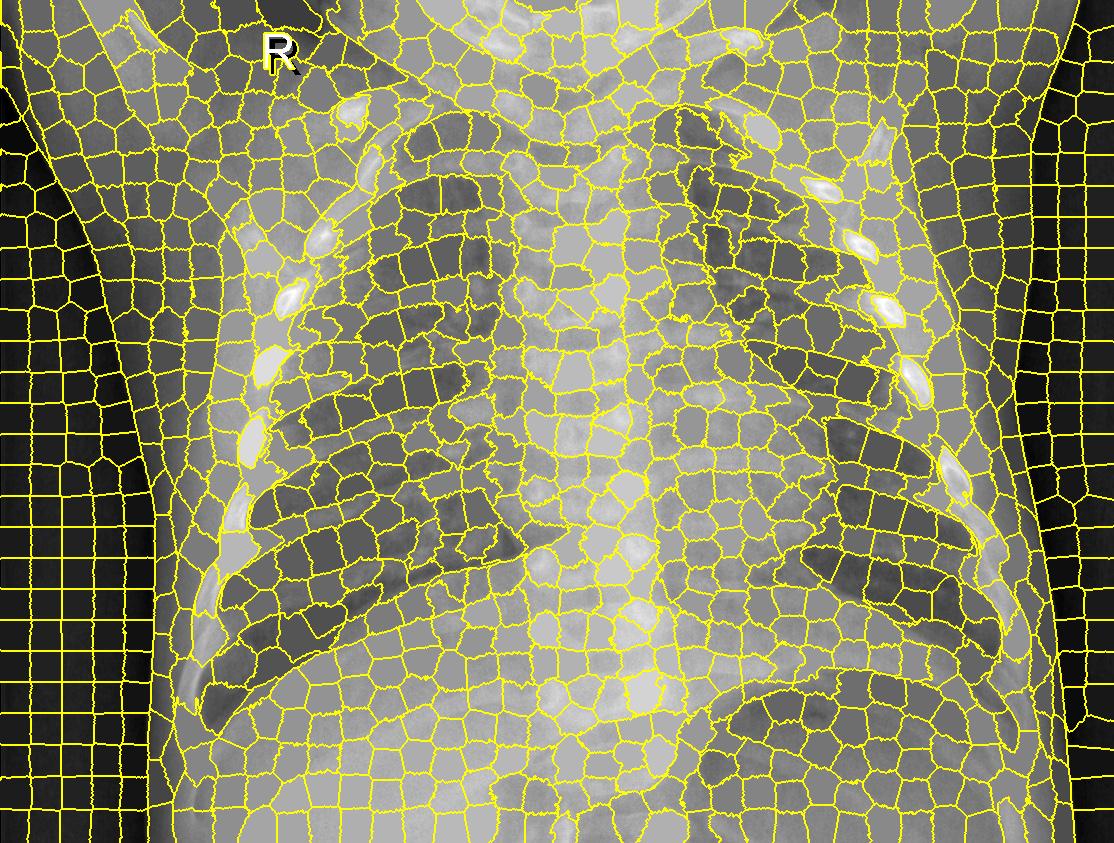}  
  \caption{SuperpixelGridMean.}
  \label{55}
\end{subfigure}
\begin{subfigure}{.18\textwidth}
  \centering
  \includegraphics[width=3cm]{IM03700001.jpeg}  
  \caption{Image $2^\prime$.}
  \label{551}
\end{subfigure}
\begin{subfigure}{.18\textwidth}
  \centering
  \includegraphics[width=3cm]{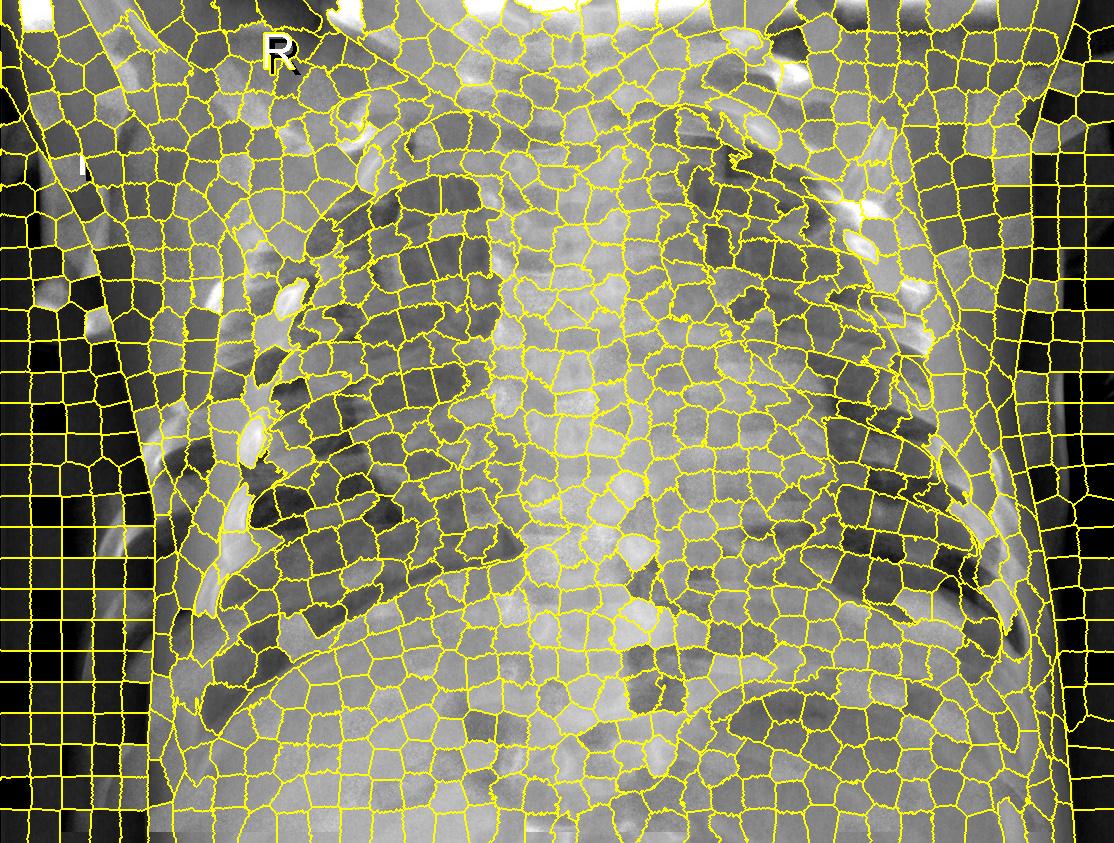}  
  \caption{SuperpixelGridMix.}
  \label{551}
\end{subfigure}\\
\begin{subfigure}{.18\textwidth}
  \centering
  \includegraphics[width=3cm]{IM03330001.jpeg}  
  \caption{Image $1^\prime$.}
  \label{332}
\end{subfigure}
\begin{subfigure}{.18\textwidth}
  \centering
  \includegraphics[width=3cm]{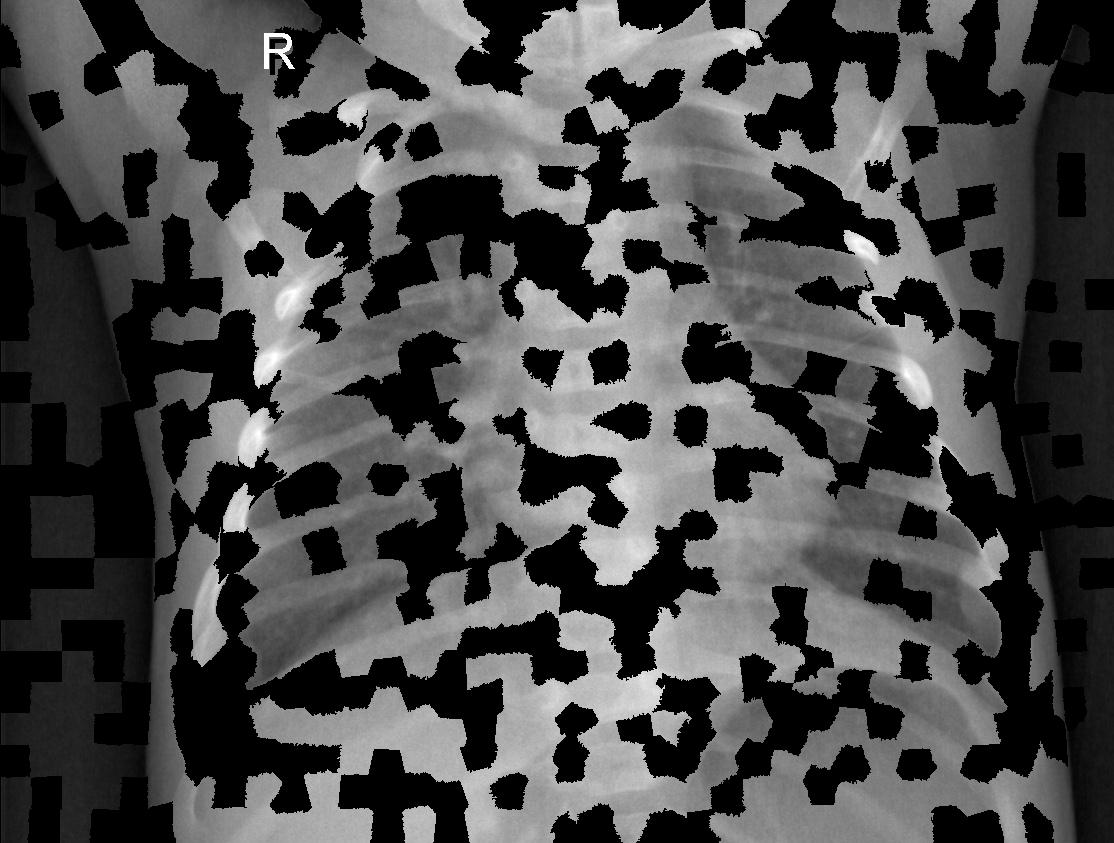}  
  \caption{SuperpixelGridCut.}
  \label{44}
\end{subfigure}
\begin{subfigure}{.18\textwidth}
  \centering
  \includegraphics[width=3cm]{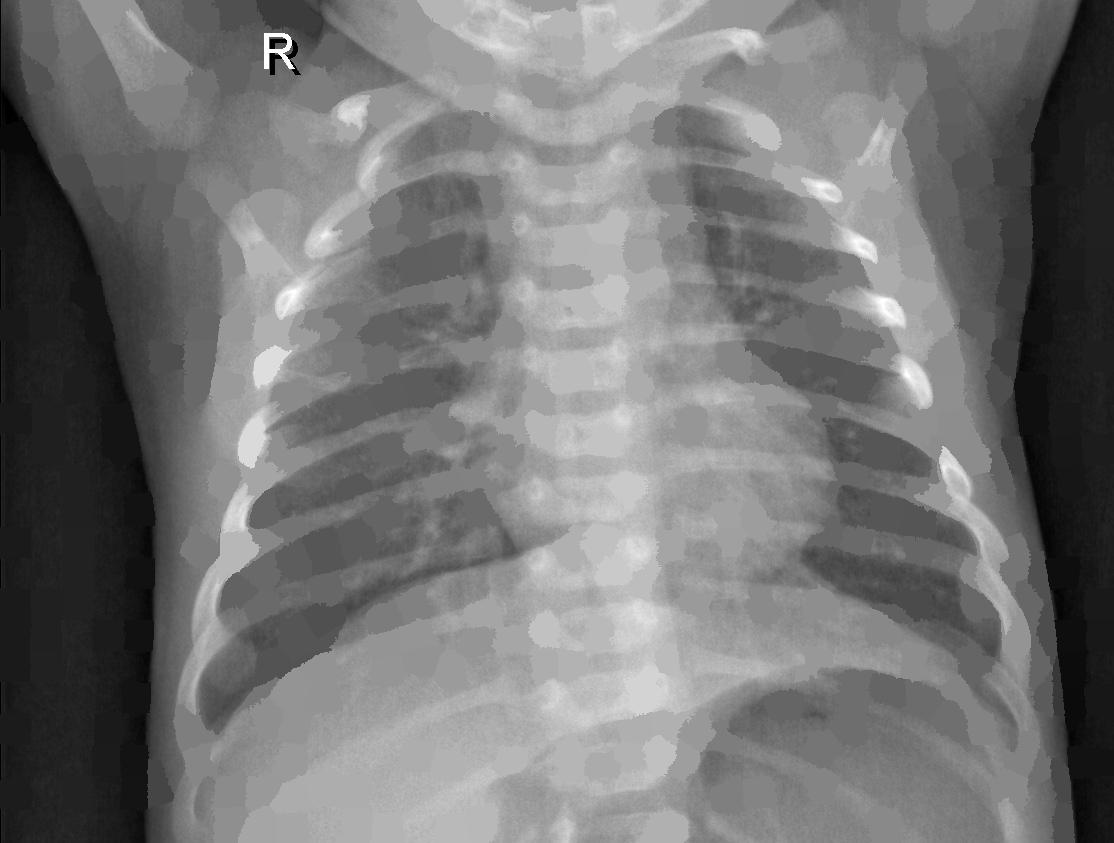}  
  \caption{SuperpixelGridMean.}
  \label{55}
\end{subfigure}
\begin{subfigure}{.18\textwidth}
  \centering
  \includegraphics[width=3cm]{IM03700001.jpeg}  
  \caption{Image $2^\prime$.}
  \label{551}
\end{subfigure}
\begin{subfigure}{.18\textwidth}
  \centering
  \includegraphics[width=3cm]{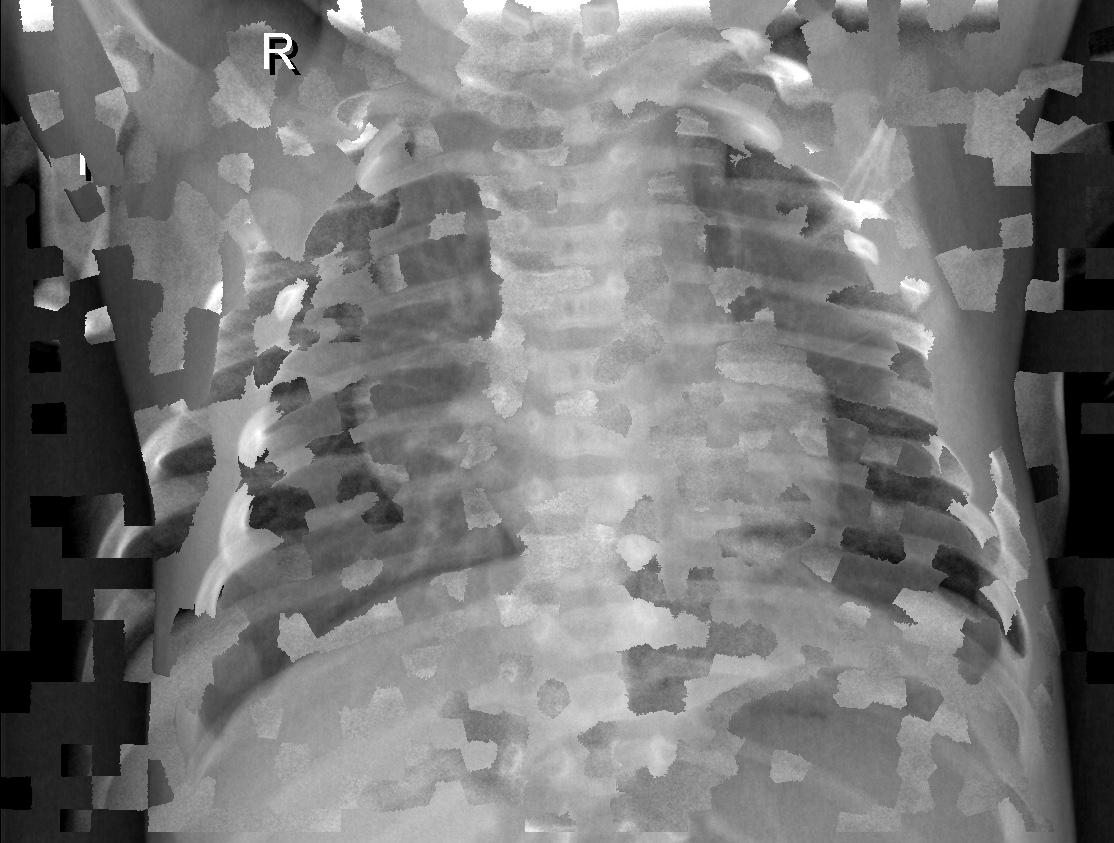}  
  \caption{SuperpixelGridMix.}
  \label{551}
\end{subfigure}
\caption{Augmentations generated by applying the proposed \textit{SuperpixelGrixCut}, \textit{SuperpixelGridMean} and \textit{SuperpixelGridMix} over X-Ray images. Top rows and bottom rows show our results with two selected superpixel quantities, $q=200$ and $q=1000$, respectively. The selected ratio of processed superpixels is $r=0.4$.}
\label{experiment2}
\end{figure*}

\subsection{Analysis of Generated SuperpixelGridMasks}

Training image sets of previously described datasets have been augmented using \textit{SuperpixelGridCut}, \textit{SuperpixelGridMean} and \textit{SuperpixelGridMix}. The SLIC Superpixel segmentation method has been used. Low and high quantities of generated superpixels as well as low and high contiguity level associated with the processed superpixels have been experimented. A sample of generated \textit{SuperpixelGridMask} images are shown in Figures \ref{teaser}, \ref{experiment1} and \ref{experiment2} with varied respective datasets.

In particular, Figure \ref{experiment1} shows augmentations generated by applying the proposed \textit{SuperpixelGrixCut}, \textit{SuperpixelGridMean} and \textit{SuperpixelGridMix} over urban and semi-urban scene images. Top rows and bottom rows show our results with two selected superpixel quantities, $q=200$ and $q=500$, respectively. The selected ratio of processed superpixels is $r=0.4$. We can see that the images contain foreground, object and background parts which mainly are road, car and forest or building facade textures.

We can observe that the superpixel segmentation has permitted to segment object parts (on the contrary of quadrilateral segmentation operated by cutting techniques). For instance, in the first row, car parts (a wheel, part of doors, driver's window) are distinguishable when using the \textit{SuperpixelGridCut}. The \textit{SuperpixelGridMean} which sparsely averages superpixels acts on the images like a slight blurring of object parts. It looks like low variations of textures and shapes of object parts at local positions (see e.g. the deformed tree trunks or road lines). Besides, we can observe that the \textit{SuperpixelGridMix} which exploited two images having different backgrounds (forest and building facade, respectively) has created an image which looks like having a new background appearing as a hedge of shrubs at the bottom of the building facade. We emphasize that the superpixel grid can be more or less regular when the superpixels are small and located in homogeneous areas (e.g. see road superpixels at the bottom row). 

In particular, Figure \ref{experiment2} shows augmentations generated by applying the proposed \textit{SuperpixelGrixCut}, \textit{SuperpixelGridMean} and \textit{SuperpixelGridMix} over X-Ray images. Top rows and bottom rows show our results with two selected superpixel quantitites, $q=200$ and $q=1000$, respectively. The selected ratio of processed superpixels is $r=0.4$. We can see that the X-Ray superpixels contain well delimited object parts (e.g. lungs, ribs). Proposed SuperpixelGridMasks have permitted to generate object-related variants of Chest X-Ray images towards further experiments.

\begin{figure}[t]
\centering
   \begin{subfigure}{0.48\linewidth}
   \centering
   \includegraphics[width=\linewidth]{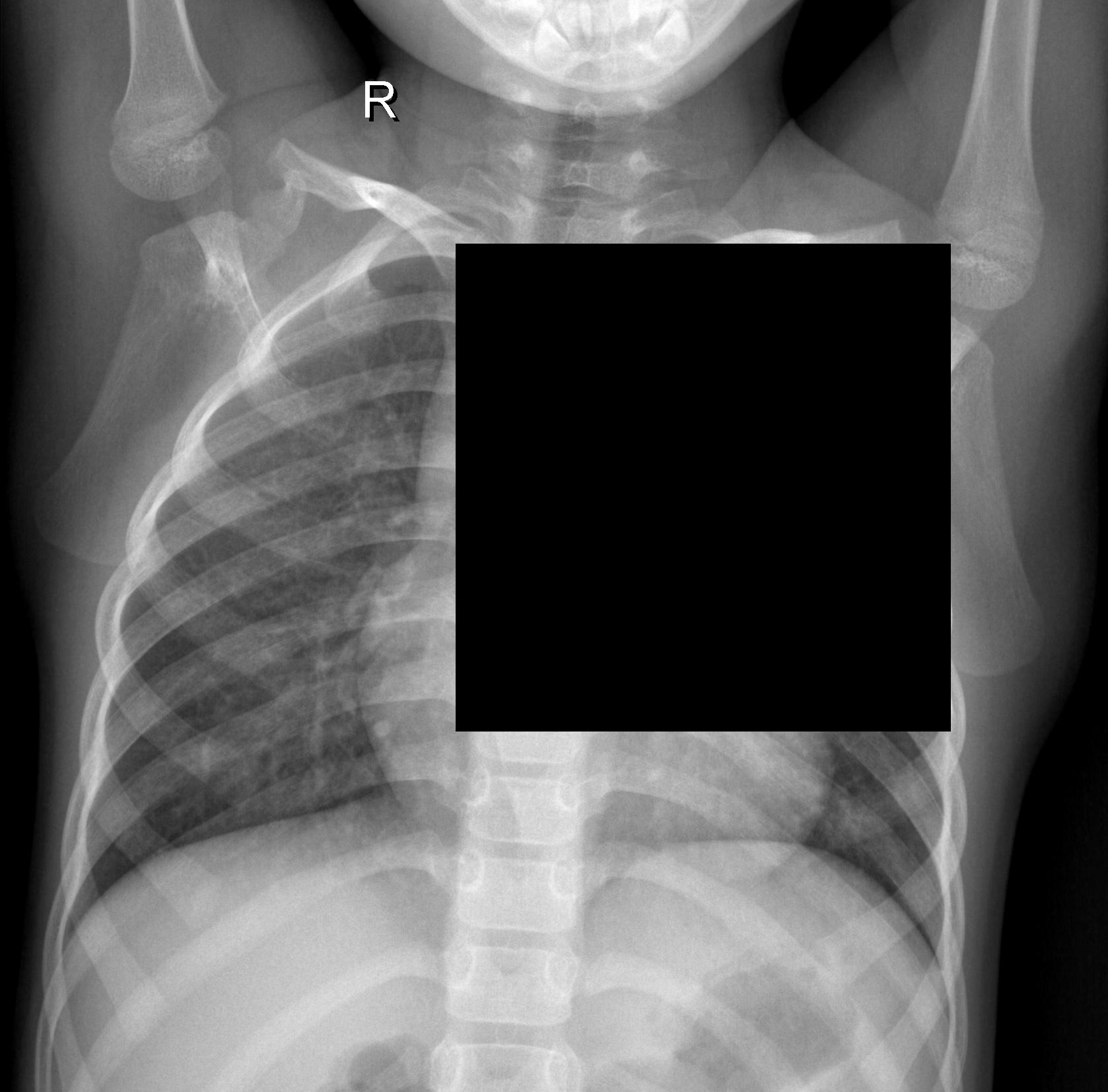}
   \caption{Sample of a CutOut.}
   \label{tsnea} 
\end{subfigure}
\hfill
\begin{subfigure}{0.48\linewidth}
   \centering
   \includegraphics[width=\linewidth]{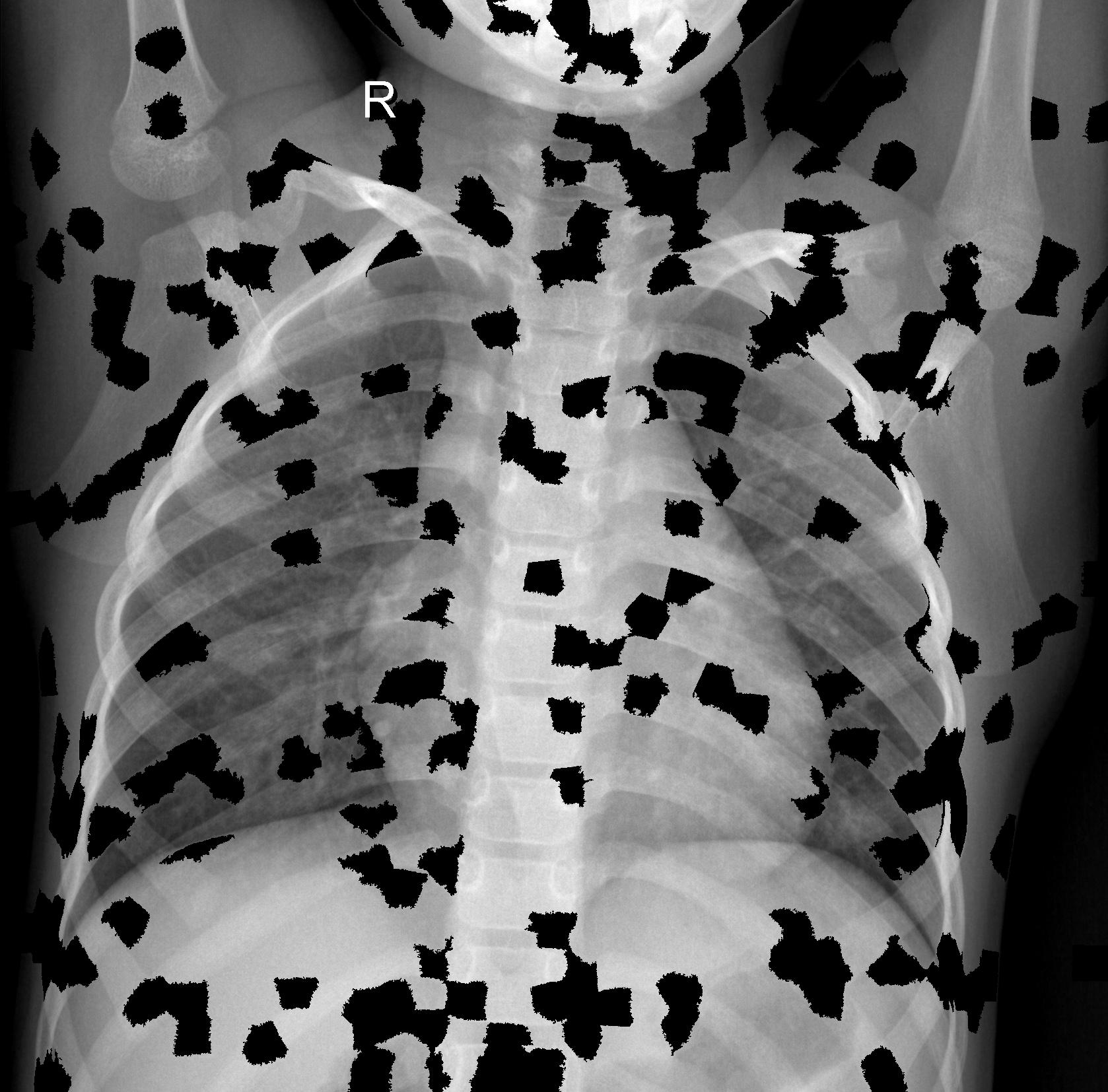}
   \caption{Sample of a SuperpixelGridCut.}
   \label{tsneb}
\end{subfigure}
\\
\begin{subfigure}[H]{\linewidth}
   \centering
   \includegraphics[width=1\linewidth]{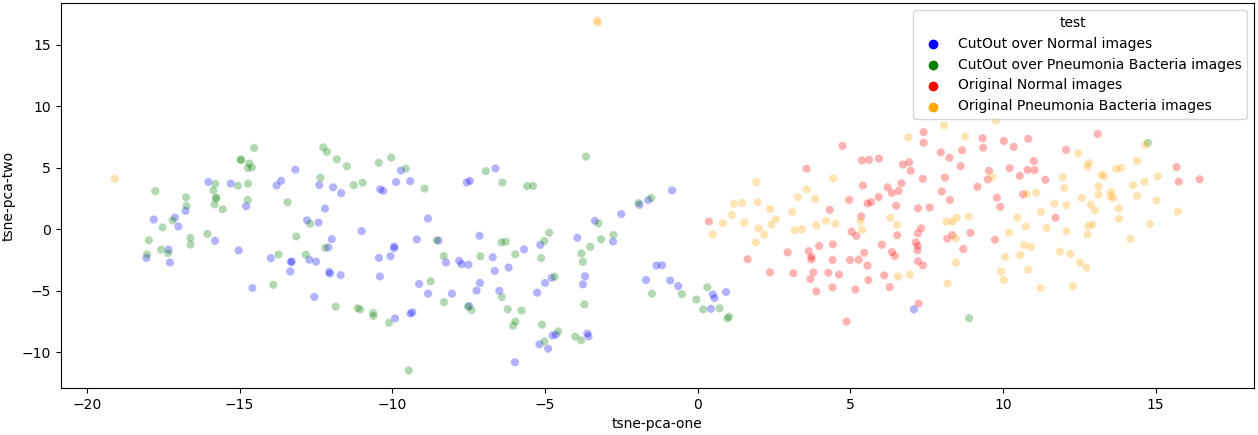}
   \caption{No augmentation Vs. CutOut augmentation.}
   \label{tsnec}
\end{subfigure}
\\
\begin{subfigure}[H]{\linewidth}
   \centering
   \includegraphics[width=1\linewidth]{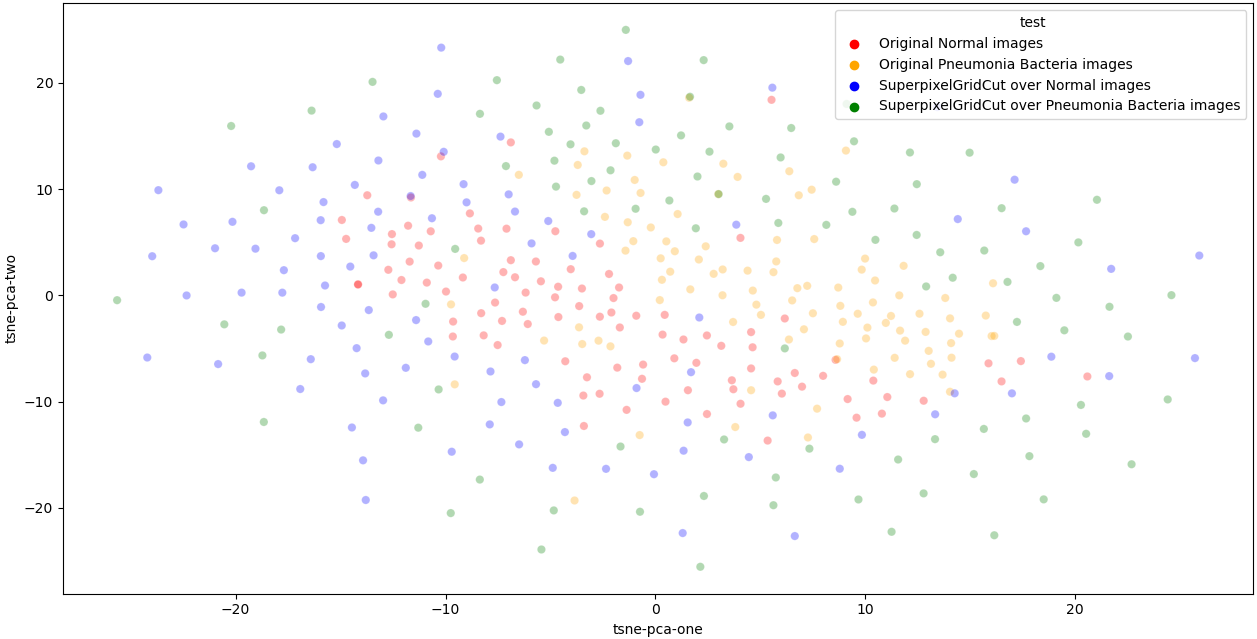}
   \caption{No augmentation Vs. SuperpixelGridCut augmentation.}
   \label{tsned}
\end{subfigure}
\centering
\caption{\label{tsne}Subfigures \ref{tsnea} and \ref{tsneb} show images from the \textit{Chest X-Ray Images} when augmented by CutOut and SuperpixelGridCut with a similar masking ratio of approximately 0.2. Subfigures \ref{tsnec} and \ref{tsned} show the distribution of points (images) generated via t-SNE by using 100 \textit{Normal} and 100 \textit{Bacterial pneumonia} images without augmentation (Originals) compared to their generated CutOut and SuperpixelGridCut augmentations, respectively.}
\end{figure}

Besides, we emphasize that \textit{SuperpixelGridCut} is the most penalizing data augmentation method by comparison to \textit{SuperpixelGridMean} or \textit{SuperpixelGridMix}. Indeed, \textit{SuperpixelGridCut} occludes a set of superpixels for considered original images with black masks whereas \textit{SuperpixelGridMean} or \textit{SuperpixelGridMix}, either averaged superpixels or mixed superpixels, respectively. 

In Figure \ref{tsne}, we analyzed the behavior of our \textit{SuperpixelGridCut} augmentation method over original images of the \textit{Chest X-Ray Images} dataset. To this end, we compared this latter method (see a sample in Figure \ref{tsneb}) with the CutOut method which also proceeds by masking an image region by a black mask (see a sample in Figure \ref{tsnea}). For a fair comparison, the ratio of masking of both is approximately equal to 0.2. For comparing, samples of original training images and corresponding augmented images have been passed through a data dimensionality reduction method. In particular, PCA (Principal Component Analysis) and t-SNE (t-distributed stochastic neighbor embedding) \cite{tsne} have been applied to generate a distribution of points towards visualization and analysis of image samples (400 images used for each generated view).

In Figure \ref{tsnec}, we can observe that two dissociated clusters of compact points are formed; one corresponds to the not augmented images (normal and pneumonia), the other one corresponds to the images augmented by using CutOut. The Figure \ref{tsned} shows a single cluster of points with not augmented images regrouped at its center and images augmented by \textit{SuperpixelGridCut} homogeneously distributed in its periphery. Although CutOut and \textit{SuperpixelGridCut} representations mask in black an approximately similar proportion for one considered image, their applications to a same set of original images conduct to the generation of augmented datasets with different distribution characteristics.

\subsection{Analysis of Classification Results}

Table \ref{teasertab} reports top-1 accuracies of our classification results using VGG19 over a PASCAL VOC dataset and the Chest X-Ray Images dataset.

\begin{table}[t]
\centering
{
\resizebox{\columnwidth}{!}{%
\begin{tabular}[t]{ccc}
\toprule
\textbf{Augmentation} & \textbf{Parameters} & \textbf{Accuracy} (\%)\\
\midrule
 Baseline & - & \textbf{57.81}\\
\midrule
Horizontal flip \cite{info11020125} & - & 62.50\\
Adjust brightness \cite{info11020125} & delta=0.1 & 57.81\\
CutOut \cite{CutOut}& r=0.2 & 65.63\\
CutOut \cite{CutOut}& r=0.4 & 65.63\\
SuperpixelGridCut (Ours) & (q=100, $r=0.2$) & 64.06\\
SuperpixelGridMean (Ours) & (q=1000, $r=0.4$) & \textbf{75.00} \textbf{(+17.19)}\\
\midrule
CutMix \cite{CutMix} & r=0.2 & 67.19\\
CutMix \cite{CutMix} & r=0.4 & 60.94\\
SuperpixelGridMix (Ours) & (q=100, $r=0.2$) & \textbf{75.00} \textbf{(+17.19)}\\
\bottomrule
\end{tabular}}
}
\caption{Classification results of our proposed data augmentation methods and diverse ones obtained using VGG19 over a \textit{PASCAL VOC}{\protect\footnotemark[1]} dataset using varied parametrization.}
\label{table1}
\end{table}%



More precisely, Table \ref{table1} presents classification results of our proposed data augmentation methods and diverse ones obtained using VGG19 over a \textit{PASCAL VOC} dataset using varied parametrization. The first row shows a baseline accuracy which is equal to 57.81\%. The augmentation methods are divided into two parts, the methods proceeding using a single image and the methods mixing two images. In the first part, we remark that \textit{SuperpixelGridMean} provides the best accuracy of 75.00\%. CutOut and \textit{SuperpixelGridCut} obtain close accuracies of 65.63\% and 64.06\%, respectively. Horizontal flip provides an accuracy of 62.50\% while the Adjust brightness equals the baseline accuracy. In the second part, CutMix reaches an accuracy of 67.19\% and \textit{SuperpixelGridMix} provides an accuracy of 75.00\%. Hence, \textit{SuperpixelGridMix} as well as \textit{SuperpixelGridMean} obtain the best results amongst the considered augmentation methods and $+17.19\%$ compared to the baseline accuracy of 57.81\%.

\begin{table}[t]
\centering
{
\resizebox{\columnwidth}{!}{%
\begin{tabular}[t]{ccc}
\toprule
\textbf{Augmentation} & \textbf{Parameters} & \textbf{Accuracy} (\%)\\
\midrule
Baseline & - & \textbf{78.68}\\
\midrule
Horizontal flip \cite{info11020125} & - & 79.32\\
Random brightness \cite{info11020125} & range=[0.5,2.0] & 82.37\\
CutOut \cite{CutOut} & r=0.2 & 80.60\\
CutOut \cite{CutOut} & r=0.4 & 80.28\\
SuperpixelGridCut (Ours) & (q=1000, $r=0.2$) & 80.44\\
SuperpixelGridMean (Ours) & (q=1000, $r=0.2$) & \textbf{83.65} \textbf{(+4.97)}\\
\midrule
CutMix \cite{CutMix} & r=0.2 & 85.09\\
CutMix \cite{CutMix} & r=0.4 & 79.80\\
SuperpixelGridMix (Ours) & (q=1000, $r=0.4$) & \textbf{88.14} \textbf{(+9.46)}\\
\bottomrule
\end{tabular}}
}
\caption{Classification results of our proposed data augmentation methods and diverse ones obtained using VGG19 over the \textit{Chest X-Ray Images}{\protect\footnotemark[2]} (Kaggle) dataset using varied parametrization.}
\label{table2}
\end{table}%

Table \ref{table2} presents classification results of our proposed data augmentation methods and diverse ones obtained using VGG19 over the \textit{Chest X-Ray Images} dataset using varied parametrization. More precisely, the first row shows a baseline accuracy which is equal to 78.68\%. The augmentation methods are divided into two parts, the methods proceeding using a single image and the methods mixing two images. In the first part, we remark that \textit{SuperpixelGridMean} provides the best accuracy of 83.65\%. The random brightness reaches an accuracy of 82.37\%. CutOut (2 parametrizations) and \textit{SuperpixelGridCut} obtain close accuracies of 80.60\%, 80.28\% and 80.44\%, respectively. In the second part, CutMix provides to the best an accuracy of 85.09\% and \textit{SuperpixelGridMix} reaches an accuracy of 88.14\%. This latter obtains the best results amongst the considered augmentation methods and $+9.46\%$ compared to the baseline accuracy of 78.68\%.

We observe that our methods can overpass baseline results. Our experiments show that \textit{SuperpixelGridCut} and CutOut can provide close performances. In this study, \textit{SuperpixelGridMean} outperforms CutOut and \textit{SuperpixelGridMix} is the most efficient method amongst all the considered data augmentation methods.  

\footnotetext[1]{A \textit{PASCAL VOC} dataset: \url{http://host.robots.ox.ac.uk/pascal/VOC/databases.html\#VOC2005_1}}
\footnotetext[2]{Dataset \textit{Chest X-Ray Images}: \url{https://www.kaggle.com/datasets/paultimothymooney/chest-xray-pneumonia}}

\section{Conclusion}

A new efficient and flexible object-related data augmentation approach has been presented. Proposed \textit{SuperpixelGridCut}, \textit{SuperpixelGridMean} and \textit{SuperpixelGridMix} variants act limiting the inclusion of linear, right-angled and regular noises in the considered image analysis networks. A bunch of CutOut and CutMix-related data augmentation methods that currently exploit linear structures could easily adopt our approach in order to adapt their existing models towards targeting better performances. The source codes of each proposed method has publicly been made available online for permitting to anyone the easy augmentation of his/her image datasets and further experiments towards better performances. A SuperpixelGridMask branch is introduced in the category of dropping and fusing data augmentation methods with a high potential of generalization. 

Future works may investigate our SuperpixelGridMasks and potential variants towards improving YOLO-based detectors. Besides, since who can do more can do less, CutMean and variants could be experimented since they do not seem to be addressed in the current literature.

{\small 
\bibliographystyle{ieeetr}
\bibliography{egbib}
}

\end{document}